\documentclass{article}

\usepackage{graphicx}
\usepackage{multirow}
\usepackage{colortbl}
\usepackage[ruled]{algorithm2e}  

\usepackage[preprint]{neurips_2022}

\usepackage[utf8]{inputenc} 
\usepackage[T1]{fontenc}    
\usepackage{hyperref}       
\usepackage{url}            
\usepackage{booktabs}       
\usepackage{amsfonts}       
\usepackage{nicefrac}       
\usepackage{microtype}      
\usepackage{xcolor}         

\title{V4D: Voxel for 4D Novel View Synthesis}

\author{%
  Wanshui Gan \\
  The University of Tokyo, RIKEN AIP\\
  \texttt{wanshuigan@gmail.com} \\
   \And
   Hongbin Xu \\
   South China University of Technology, Alibaba Group \\
   \texttt{xuhongbin.xhb@alibaba-inc.com} \\
   \AND
   Yi Huang \\
   Shenzhen Institute of Advanced Technology,\\ Chinese Academy of Sciences\\
   \texttt{yi.huang@siat.ac.cn} \\
   \And
   Shifeng Chen \\
   Shenzhen Institute of Advanced Technology,\\ Chinese Academy of Sciences\\
   \texttt{shifeng.chen@siat.ac.cn} \\
   \And
   Naoto Yokoya \\
   The University of Tokyo, RIKEN AIP \\
   \texttt{yokoya@k.u-tokyo.ac.jp} \\
}

\begin{document}

\maketitle

\begin{abstract}
  Neural radiance fields have made a remarkable breakthrough in the novel view synthesis task at the 3D static scene. However, for the 4D circumstance (e.g., dynamic scene), the performance of the existing method is still limited by the capacity of the neural network, typically in a multilayer perceptron network (MLP). In this paper, we utilize 3D Voxel to model the 4D neural radiance field, short as V4D, where the 3D voxel has two formats. The first one is to regularly model the 3D space and then use the sampled local 3D feature with the time index to model the density field and the texture field by a tiny MLP. The second one is in look-up tables (LUTs) format that is for the pixel-level refinement, where the pseudo-surface produced by the volume rendering is utilized as the guidance information to learn a 2D pixel-level refinement mapping. The proposed LUTs-based refinement module achieves the performance gain with little computational cost and could serve as the plug-and-play module in the novel view synthesis task. Moreover, we propose a more effective conditional positional encoding toward the 4D data that achieves performance gain with negligible computational burdens. Extensive experiments demonstrate that the proposed method achieves state-of-the-art performance at a low computational cost. The relevant code will be available in \href{https://github.com/GANWANSHUI/V4D}{https://github.com/GANWANSHUI/V4D}.
\end{abstract}

\section{Introduction}

The novel view synthesis task could offer an immersive experience in applications such as augmented reality, virtual reality, games, and the movie industry, which has been attracting more and more attention in recent years. The differentiable volume rendering technique has significantly boosted the performance in novel synthesis tasks, where the representative one in these two years should be the Neural Radiance Fields (NeRF) proposed in \cite{nerf}. In NeRF, the relative soft geometry representation via volume rendering could make the geometry learn more efficiently, especially under multi-view constraints, which is a key distinguishing characteristic compared with the geometry representation with decision boundaries such as the occupancy field \cite{mescheder2019occupancy} and signed distance field \cite{yariv2021volume}. The novel view synthesis in static scenes has been well studied, such as for large scenes \cite{xiangli2021citynerf, tancik2022block, rematas2021urban}, low computational cost \cite{liu2020neural, sun2021direct, plenoctrees, yu2021plenoxels}, relaxing the camera pose or the number of posed images \cite{yu2021pixelnerf, chen2021mvsnerf}, and better geometry representation \cite{wang2021neus, oechsle2021unisurf, yariv2021volume}. In this paper, we focus on the novel view synthesis in dynamic scenes, particularly under a single-view video setting, which is much more challenging due to the lack of efficient multi-view constraints.

For the novel view synthesis in the dynamic scenes, one of the problem settings is under multi-view video and this setting could usually produce better results thanks to the existed multi-view constraints at the same moment. However, capturing the multi-view video relies on the support from professional equipment, which asks for the multi-view camera rigs. These laborious and expensive setting makes people tend to explore the single-view video in a dynamic scene, even though this is an ill-posed problem lacking the multi-view constraints. To alleviate the ill-posed problem, previous works introduce the constraint information from the third part module, such as the optical flow between the adjacent video frames and the monocular depth information for the geometry consistency constraints \cite{xian2021space, li2021neural, gao2021dynamic}. Another research line is to predict the canonical space of the dynamic scene and then to model the neural radiance field at the canonical space \cite{park2021nerfies, pumarola2021d, tretschk2021non}. However, the mentioned methods only used the MLPs to model the neural radiance field, and we argue that the representation ability would be limited due to the neural network's capacity, especially under the lack of multi-view constraints. To handle this problem, we propose the 3D voxel-based architecture (V4D) to model the 4D dynamic neural radiance field. The overview of the proposed architecture is illustrated in Figure \ref{fig:Overview of v4d}.

Specifically, we model the neural density field and the texture field separately under the hybrid network structure, where the sampled feature in the voxel grid is combined with the time index and then pass to an MLP for the density and the RGB texture information. However, the hybrid network structure is easy to over-fit to the training view due to the lack of multi-view constraints. We find that the total variation loss on the voxel grid could effectively prevent this problem and maintain proper geometry learning. Although we can achieve significant performance gain with the total variation constraints on the voxel grids, the high-frequency detail is still not well delineated and tend to be over smooth on the surface. Therefore, we introduce a conditional positional encoding module to recap the high-frequency details. Moreover, after the volume rendering, we further design the pixel-level refinement module for a better color representation. Note that the proposed two modules only consume little computational cost and achieve performance improvement. At last, the proposed V4D is compared with the single view video datasets proposed in \cite{pumarola2021d} and \cite{park2021hypernerf}. The extensive experiments demonstrate the superiority of the proposed method.

In summary, the main contributions of this work are as follows:
\begin{itemize}
 \item We propose the method V4D, a simple yet effective and efficient framework, for 4D novel view synthesis with the 3D voxel, which directly models the 4D neural radiance field without the need for canonical space.  
 
 \item To address the over-smooth problem caused by the total variation loss, we propose the conditional positional encoding module and pixel-level refinement module, which are verified to be effective in improving the performance, whereas the pixel-level refinement module implemented by the look-up tables could be regarded as the plug-and-play module in the novel view synthesis task. 
 
  \item The proposed V4D achieves state-of-the-art performance in synthesis dataset \cite{pumarola2021d} and competitive performance in real scenes dataset \cite{park2021hypernerf}.

 \end{itemize}

\section{Related works}

\subsection{Neural radiance field representation}
Multilayer perceptrons (MLPs) are widely used to construct the neural radiance field in a compact network, but researchers also recognize the shortcoming that the global optimization of the whole MLPs is time-consuming. Therefore, for a fast rendering speed, \cite{plenoctrees} proposes NeRF-SH by pre-tabulating the NeRF into a PlenOctree and factorizing the appearance spherical harmonic representation. \cite{wang2022fourier} further proposes plenoptic voxels to represent the 4D dynamic scene with spherical harmonics under the multi-view video setting. \cite{liu2020neural, sun2021direct, hedman2021baking, peng2020convolutional} are more related to this work that uses the hybrid representation that learns a voxel and MLPs to model the radiance field and the view-dependent effect simultaneously, achieving fast training and rendering speed. Some very recent methods also share a similar concept of the dual radiance field. \cite{shao2021doublefield} and \cite{yao2021dd} propose Doublefield and double diffusion based neural radiance field for high-fidelity human reconstruction and rendering, which is towards static multi-view reconstruction. Different from the above methods, the proposed V4D is for the 4D scene representation in a single-view video setting, which is more challenging.  

\subsection{4D representation}
Apart from the 3D scene representation, it is natural to consider the neural radiance field for the 4D situation and it could be a single 3D scene with a time dimension, a dynamic scene, or just multiple independent 3D scenes. For multiple independent 3D scenes, there has been a limited number of methods explored in this field. \cite{liu2020neural} has revealed the ability to memorize multiple independent 3D scenes with its proposed method, but it needs to learn an independent voxel embedding for each scene and only share the same MLPs to predict density and color. For the dynamic scenes, \cite{Niemeyer_2019_ICCV} proposes occupancy flow, a spatio-temporal representation of time-varying 3D geometry with implicit correspondences, which could be built from the images or point clouds. More recently, researchers modeled the dynamic scene with the neural radiance field, which could offer a more immersive visual experience. There are two main methods. The first is to learn the 4D radiance field, which conditions the radiance field with 4D vector such as the 3D position plus with time stamp \cite{xian2021space, li2021neural, gao2021dynamic}. The second is by learning an intermediate time-invariant 3D representation or canonical space \cite{park2021nerfies, pumarola2021d, tretschk2021non}. However, the mentioned methods require the third part supervision signal (e.g., optical flow, depth information) to learn the 4D radiance field or achieve unsatisfying performance and high computational cost. In contrast, the proposed V4D does not need the additional supervision signal apart from the collected sequence of posed images, and it achieves the superior 4D scene representation ability with a much lower computational resource requirement.

\subsection{Look-up tables and concurrent work}

The proposed pixel-level refinement module is related to the works in the image enhancement task \cite{lut1, lut2}. \cite{lut1} proposes image-adaptive 3D LUTs for real-time image enhancement and \cite{lut2} considers the adaptive 3D LUTs with the global scenario and local spatial information, which could get better results. Different from them, we treat the 3D LUTs as the refinement module in our 4D novel view synthesis task with dedicated design. To the best of our knowledge, this work is the first work to explore the LUTs refinement in novel synthesis tasks. Besides, as this work is finished, we find two concurrent works \cite{TiNeuVox, NDVG} also share the idea model the neural radiance field by the voxel. For details, TiNeuVox \cite{TiNeuVox} proposes a tiny coordinate deformation network to model coarse motion trajectories. NVDG \cite{NDVG} uses a deformation grid to store 3D dynamic features, and a light-weight MLP maps a 3D point in observation space to the canonical space using the interpolated features. Both of them use the canonical space to model the neural radiance field. Different from them, we directly do the 4D attributes mapping to the density and color without canonical space and achieves better result as introduced as follows.

\begin{figure*}
\setlength{\belowcaptionskip}{-0.3cm}
\centering
\includegraphics[width=1.1\textwidth]{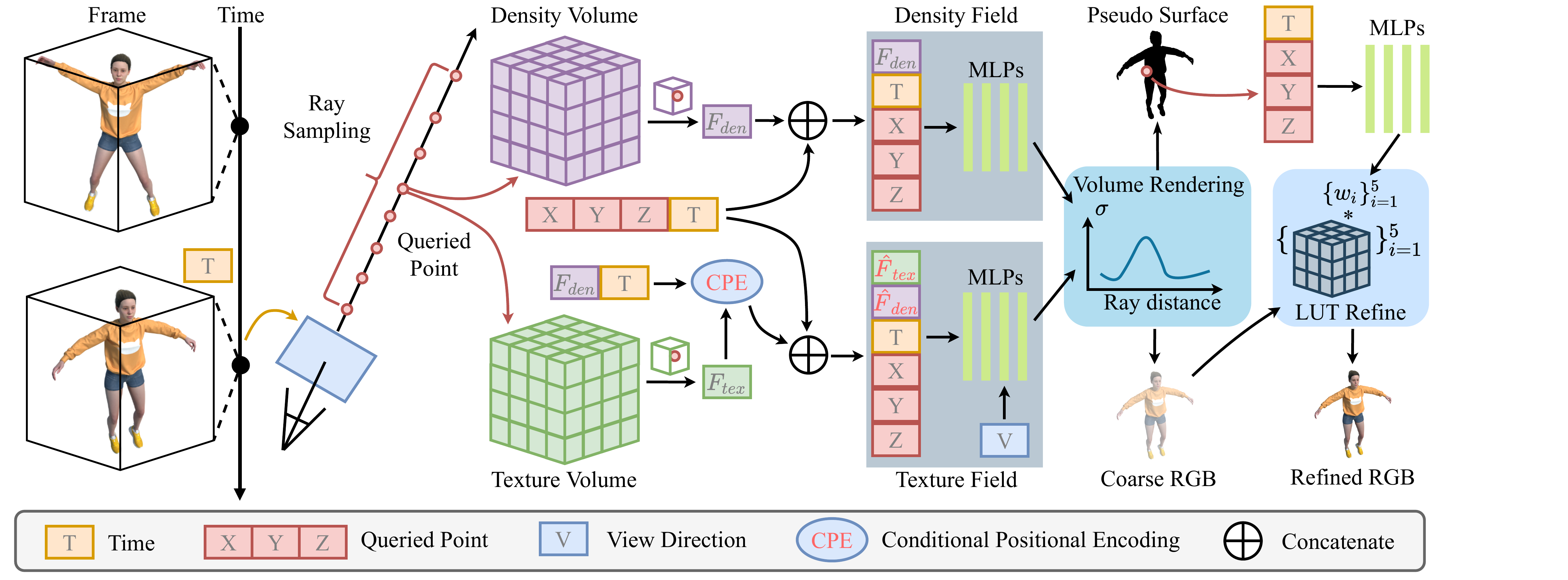}
\caption{\textbf{Overview of the voxel for 4D novel view synthesis (V4D).} Given a single-view video clip, we present a voxel-based method for 4D novel view synthesis. The detailed introduction of the conditional positional encoding (CPE) and the look-up tables refinement module (LUT refine) are placed in Section \ref{v4d}.}
\label{fig:Overview of v4d}
\end{figure*}

\section{Method}

\subsection{Preliminaries}

In this paper, we discuss the voxel-based architecture for 4D novel view synthesis. By extending the 3D situation to 4D, we need to learn a proper mapping function to map the 6D attributes $(x,y,z,t,\theta, \phi)$ into the volume density $\sigma$ and the RGB color $c$, where $(x,y,z)$ is the 3D spatial location of the each sample point, $t$ is the time index at that moment, and $(\theta, \phi )$ is the 2D view direction for modeling the view dependent effect. We define the mapping function as $M$: $(x,y,z,t,\theta, \phi) \rightarrow (\sigma, c)$. Following the approximate volume rendering in NeRF \cite{nerf}, we can learn the mapping function by supervising the estimated RGB color $\hat{c}$. The estimated RGB color of each pixel in image plane could be obtained by the equation (\ref{rendering}) : 
\begin{equation}
    \hat{c} = \sum_{i=1}^{N}{T_{i}(1-exp({\sigma _{i}}{\delta _{i}}  ) c_{i} ) }, 
     \label{rendering}
\end{equation} 
where ${T_{i} = exp\left ( -\sum_{j=1}^{i-1} {\sigma _{j}}{\delta _{j}}  \right ) }, {\delta _{i}} = t_{i+1} - t_{i}$, $i$ is the sampled point along the ray, and $N$ is the number of the sampled points.
 
The key component is to design the effective neural network as the mapping function, and we present our method as follows.

\subsection{Voxel for 4D novel view synthesis} \label{v4d}

\textbf{Network design}
The previous work has revealed the advantage of using the voxel as the backbone for the 3D novel view synthesis, which could be computationally efficient and have higher accuracy compared with the MLP-based methods. We propose V4D, as illustrated in Figure \ref{fig:Overview of v4d}, for 4D novel view synthesis. Since it would cost huge memory storage if we adopt the 4D voxel format, we initialize two 3D voxels with dimensions $160\times160\times160\times12$, the density volume and the texture volume, where density volume is mainly for modeling the density field and the texture volume is only for the texture field. For modeling the time dimension, we concatenate the time index $t$ with the 3D location $(x,y,z)$, view direction $(\theta, \phi )$, and the sampled feature $(F_{den}, F_{tex})$, and then feed them into the density field and texture field, respectively. We use the 5-layer MLPs to model the density field and the texture field. With the volume rendering in equation (\ref{rendering}), we can obtain the coarse RGB pixels in the image plane. As stated before, the total variation loss on the voxel grids is a key factor to prevent the neural work from over-fitting the training set, especially in the dynamic scenes. However, it is observed that the novel view result would be a bit blurred due to the over-smooth characteristics of the total variation loss, which means that the high-frequency detail is missing. Therefore, we propose the conditional positional encoding and a look-up tables refinement module to alleviate this problem as follows.


\textbf{Conditional positional encoding} The positional encoding is critical to recover the high-frequency details in the novel view synthesis task \cite{nerf}. In our proposed method, we not only do the positional encoding in the 6D attributes $(x,y,z,t,\theta, \phi)$ but also apply it to the sampled feature $(F_{den}, F_{tex})$. Besides, in the 4D situation, we further explicitly assign the phase shift to different frequencies, which is inspired by the previous work that the phase information retains the main information of the image after the Fourier transformation \cite{phaseshift}. Therefore, shifting the phase with the time index should help the neural network to learn effective feature embedding at different moments. We introduce the conditional positional encoding (CPE) defined as,
\vspace{0cm}
\begin{equation}
   \gamma (p_{v}) = \left (\sin (2^{L-1}\pi p_{v} + \frac{2\pi }{2^{L-1}\pi} t ), \cos (2^{L-1}\pi p_{v} +  \frac{2\pi }{2^{L-1}\pi} t )  \right ), 
     \label{PE}
\end{equation} 
where $p_{v}$ is the sampled feature vector at position $(x,y,z)$, $L=5$ is the frequency order, and $t$ is the time index. Note that we do not use the conditional positional encoding in the sampled feature $(F_{den})$ for the density field and only apply the CPE to the texture field, $(F_{den}, F_{tex})$ $\longrightarrow$  $(\hat{F}_{den} , \hat{F}_{tex})$. The reason is that the voxel for the density field should be initialized with zero for the correct volume rendering at the beginning of the training phase, and doing the conditional positional encoding for the density feature would break this rule that would make the learning collapse.

\begin{figure*}
    \centering
    \includegraphics[width=1\textwidth]{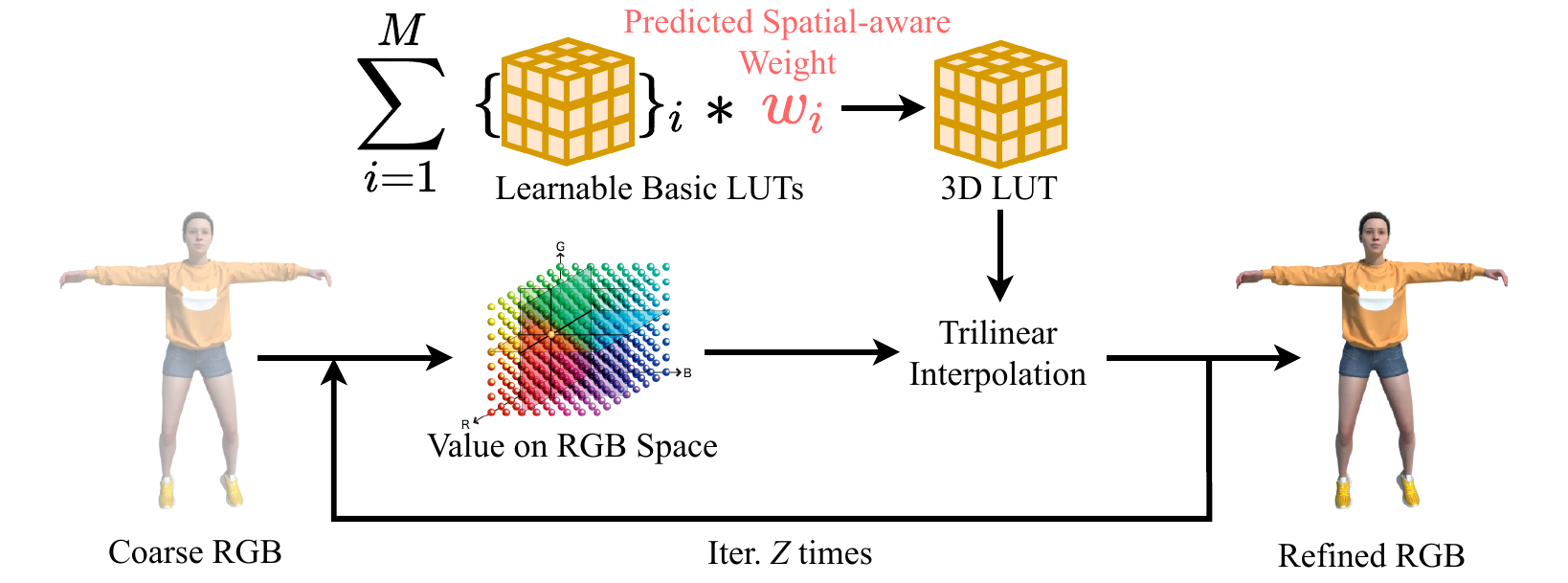}
    \caption{\textbf{LUTs refinement module.} Given the coarse RGB value as input, we learn $M=5$ basic LUTs to model a 2D pixel-level refinement mapping with guidance from the pseudo-surface. We do the recurrent iteration with $Z=3$ times for the best result. The detailed introduction is in \ref{lut}. }
    \label{fig:lut_fig}
\end{figure*}

\textbf{LUTs refinement} \label{lut}
The existing neural rendering methods directly do the 2D RGB supervision after the volume rendering and rarely consider any refinement operation. We propose voxel-based look-up tables for the pixel-level RGB refinement as shown in Figure \ref{fig:lut_fig}. Following the previous work \cite{lut1}, we also used the downsampled voxel grids with resolution $33\times33\times33$ to construct the RGB color space. We use 5 basic LUTs as the refinement units and, at the beginning of the training, one basic LUT is initialized as the identity LUT for a more stable training, and the rest is initialized as zero for a more expressive color space representation. Given the coarse RGB pixel value, we treat it as the spatial location and do the trilinear interpolation on the basic LUTs to obtain the interpolated RGB value. An important step is to combine the interpolated RGB value from the basic LUT as the final output. In the image enhancement task \cite{lut1, lut2}, the low-resolution image is used to predict the weight to combine the interpolated RGB value. However, it is unpractical in the novel view synthesis task, where such an operation loses the 3D spatial awareness. To make sure the weight of the basic LUTs is 3D spatial aware, we propose to use the pseudo-surface as input and use 10-layer MLPs to predict the weight for composing the basic LUTs. Specifically, the 3D point on the pseudo-surface could be obtained by depth information from the rendering equation, where we can slightly modify the equation (\ref{rendering}) into equation (\ref{weight}) for the depth information $\hat{d}$, 
\vspace{0cm}
\begin{equation}
    \hat{d} = \sum_{i=1}^{N}{T_{i}(1-exp({\sigma _{i}}{\delta _{i}}  ) t_{i} ) }.
     \label{weight}
\end{equation} 
Since the LUTs refinement module is designed after the volume rendering and is for the 2D pixel-level refinement, we only need a little computational cost and achieve the performance gain.

\subsection{Loss function}
 For training the proposed network, we define the loss function in equation \ref{loss}:
\begin{equation}
    Loss = w_{1} L_{rgb} + w_{2} L_{pt\_rgb} + w_{3} L_{bg} + w_{4} L_{TV}, \label{loss} \\
\end{equation}
where $L_{rgb} = \frac{1}{R} \sum_{r\in R} \left \| \hat{C} (r) - C(r) \right \|^{2}_{2}$, $R$ is the training rays in a batch size, the $\hat{C} (r)$ and $C(r)$ is the ground truth color and rendered color, respectively.\\
$L_{pt\_rgb} =  \frac{1}{R} \sum _{r \in R}\sum _{i = 1}^{K}(T_{i} a_{i}\left \| \hat{C} (r) - C(r) \right \|^{2}_{2} )$, $K$ is number queried point in texture volume. $T_{i}$ is the accumulated transmittance at the point$i$, $a_{i}$ is the probability of the termination at the point $i$. \\
$L_{bg} = -T_{K+1}log(T_{K+1}) - (1-T_{K+1})log(1-T_{K+1})$, which is to encourage the model to separate the foreground and the background \cite{sun2021direct}. $L_{TV}$ is the total variation loss works on the voxel grids, directly. We did not apply the total variation loss on the voxel of the LUTs refinement module and only apply it to the density volume and texture volume. $w_{1}, w_{2}, w_{3}, w_{4}$ are the weights of the loss function, which is set as $1.0, 0.01, 0.001, 0.1$, respectively. Note that, in $w_{4}$, we apply the exponential weight decay strategy during the training with 0.005, which could alleviate the over smooth problem caused by the total variation regularization.



\section{Experiment}

\subsection{Experiment setting} 

\textbf{Dataset} The proposed V4D is verified on the 8 synthesis datasets \cite{pumarola2021d} and 4 real scene datasets \cite{park2021hypernerf}. Besides, the proposed LUTs refinement model is also evaluated on the 3D static novel view synthesis dataset, Synthetic-NeRF \cite{nerf} and TanksTemples \cite{knapitsch2017tanks}, where we choose DVGO \cite{sun2021direct} and NeRF \cite{nerf} as the baseline method. Note that the hyperparameters of the LUTs refinement module are selected from the experiment on 3D scenes since the computational cost is much lower based on \cite{sun2021direct}, which could help our verification in a short time. Besides, since the geometry in the static scene is generally better than in single-view dynamic scenes, we could avoid this bias for a more justice evaluation.

\textbf{Implementation details} We train the neural network with 250k iterations in \cite{pumarola2021d} and 100k iterations in \cite{park2021hypernerf}. About the learning rate, the voxel in density and texture volume is 0.1 and the MLPs in the density and texture fields are 1e-3. In the LUTs refinement module, the learning rate of MLPs and voxel are both 1e-4. To optimize the neural network, we use the Adam optimizer \cite{diederik2014adam} with a batch size of 8,196 rays in the first stage and with a batch size of 4,196 rays in the second stage. Apart from the LUTs refinement module, the exponential learning rate decay is applied to the neural network with 0.1. \label{setting}


\textbf{Metrics} For quantitative evaluation, we use the following metrics to evaluate the novel view image in the testing set: (1) Peak signal-to-noise ratio (PSNR); (2) Structural similarity index measure (SSIM); (3) Perceptual quality measure LPIPS \cite{lpips}. $\uparrow$ means the value higher is better and $\downarrow$  means lower is better. For qualitative evaluation, apart from the RGB novel view image, we also show the FLIP \cite{flip, litianyue} error maps to highlight the result of the ablation study.




\subsection{Results}

\begin{figure*}
\setlength{\belowcaptionskip}{-0.3cm}
\centering
\includegraphics[width = \linewidth]{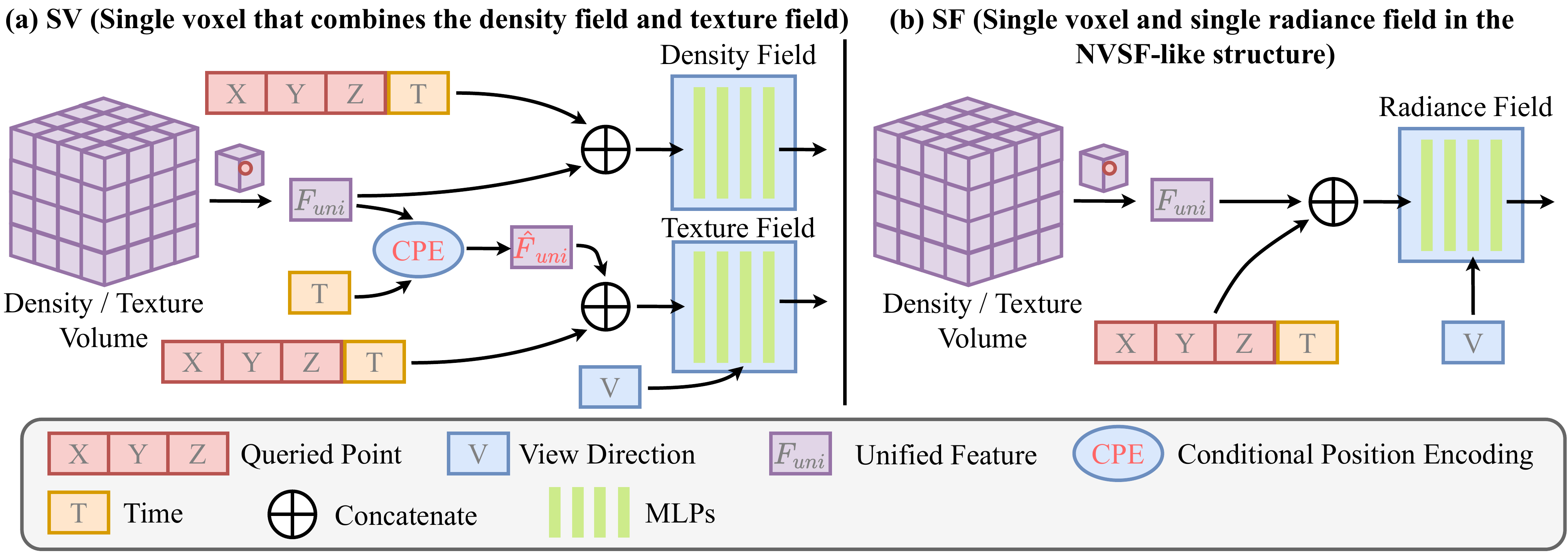}
\caption{\textbf{The variant architecture in V4D for ablation study.} For the SV, we unify the density volume and texture volume with volume size $160\times160\times160\times24$. For the SF, it is an NVSF-like structure \cite{liu2020neural} but not in the sparse voxel format. For a fair comparison, we have kept the same setting during the implementation (e.g., the width and depth of the MLPs) apart from the architecture difference illustrated above.}
\label{fig:svsf}
\end{figure*}

\begin{table*}[t]
\renewcommand\arraystretch{1.25}
\setlength{\belowcaptionskip}{0.1cm}
\centering
\caption{Quantitative comparison of the synthesis dynamic scenes under the single view setting. Our method outperforms the other methods. The training and inference time is tested on NVIDIA A100. The number with bold typeface means the best result.}
\setlength{\tabcolsep}{1mm}{}
\scalebox{0.75}{
\begin{tabular}{lcccccccccccc}

\hline
Method &  W / Time  & W / Voxel  &  W / Canonical space  &  Training (h)  & Inference (s)  &  PSNR $\uparrow$ & SSIM $\uparrow$  &  LPIPS $\downarrow$ 

\tabularnewline
\hline

NeRF \cite{nerf}  & $\times$  &  $\times$ & $\times$ & -- & -- & 19.00  & 0.87  & 0.18
\tabularnewline
T-NeRF \cite{pumarola2021d} &  \checkmark  & $\times$  & $\times$  & -- & -- &  29.51 & 0.95 & 0.08
\tabularnewline
D-NeRF \cite{pumarola2021d} & \checkmark  & $\times$  & \checkmark  & 15.9  & 15.24 &  30.50 & 0.95 & 0.07

\tabularnewline
\hline

NDVG \cite{NDVG} & \checkmark  & \checkmark   & \checkmark  & -- & -- & 31.32 & 0.97  &  0.05

\tabularnewline

TiNeuVox \cite{TiNeuVox} & \checkmark  &  \checkmark & \checkmark  & \textbf{0.35}  & 1.42 &  32.67 & 0.97 & 0.04

\tabularnewline

\hline
Ours & \checkmark & \checkmark  &  $\times$  & 6.9  & \textbf{0.48} & \textbf{33.72}  & \textbf{0.98}  & \textbf{0.02}
\tabularnewline
\hline

\end{tabular}
}

\label{t:benchmark_summarize}
\end{table*}

\begin{table}[t]
\renewcommand\arraystretch{1.25}
\setlength{\belowcaptionskip}{0.1cm}
\centering
\caption{Quantitative comparison of the real dynamic scenes under the single view setting. Our method achieves competitive performance. The number with bold typeface means the best result and the underline is the second best.}
\setlength{\tabcolsep}{1mm}{}
\scalebox{0.75}{
\begin{tabular}{lcccccccc}

\hline
Method &  PSNR $\uparrow$ & MS-SSIM $\uparrow$ 

\tabularnewline
\hline

NeRF \cite{nerf}  & 20.1 & 0.745
\tabularnewline
NV \cite{NV} &16.9 & 0.571

\tabularnewline
NSFF \cite{li2021neural}  & \textbf{26.3} & \textbf{0.916}

\tabularnewline
Nerfies \cite{park2021nerfies}  & 22.2 & 0.803

\tabularnewline

HyperNeRF \cite{park2021hypernerf}  & 22.4 & 0.814

\tabularnewline

TiNeuVox \cite{TiNeuVox}  & 24.3 & \underline{0.837}

\tabularnewline

\hline
Ours & \underline{24.8} & 0.832
\tabularnewline
\hline

\end{tabular}
}

\label{t:real_benchmark_summarize}
\end{table}

\begin{figure*}[t]
    \setlength{\belowcaptionskip}{-0.4cm}
	\centering
	\begin{minipage}[t]{0.03\linewidth}
	    \rotatebox{90}{\quad TiNeuVox \cite{TiNeuVox}}
	\end{minipage}
	\begin{minipage}[t]{0.23\linewidth}
		\centerline{\includegraphics[width=2.76cm]{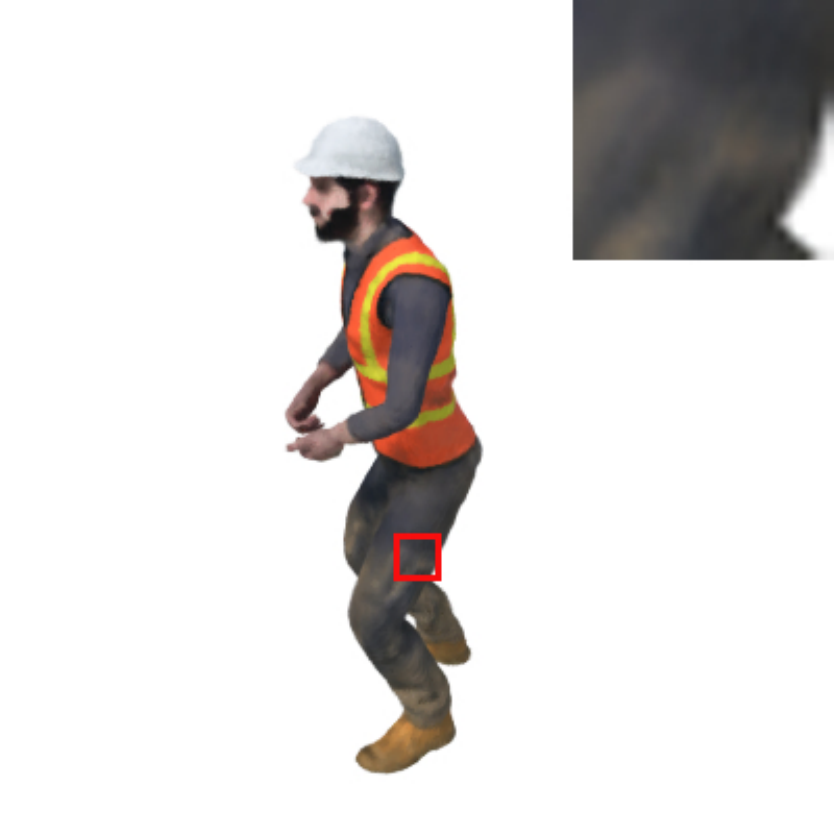}}
	\end{minipage}
	\begin{minipage}[t]{0.23\linewidth}
		\centerline{\includegraphics[width=2.76cm]{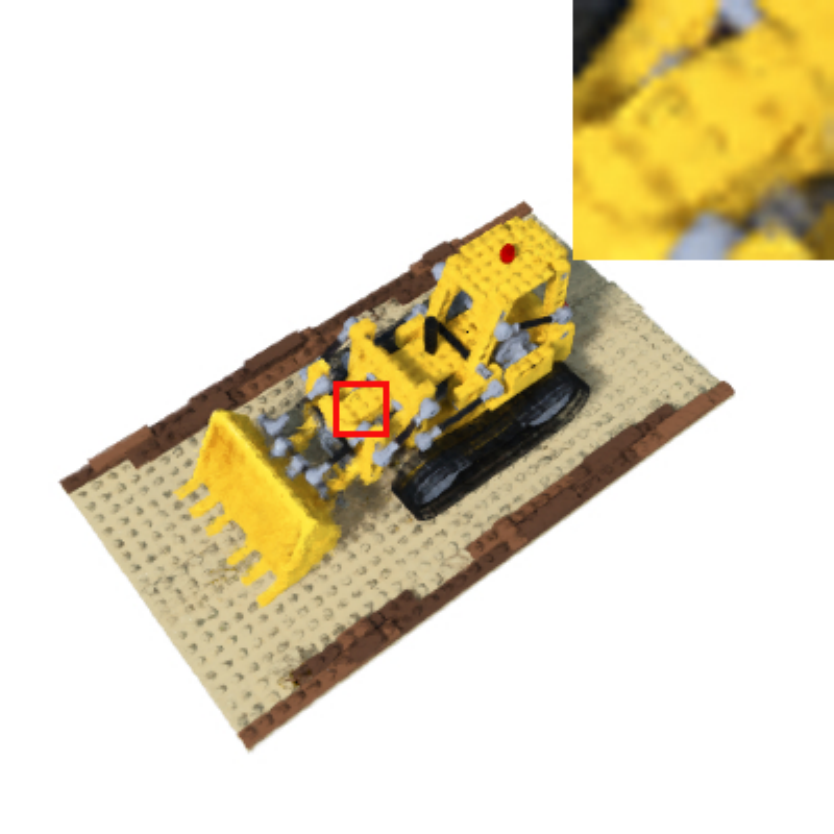}}
	\end{minipage}
	\begin{minipage}[t]{0.23\linewidth}
		\centerline{\includegraphics[width=2.76cm]{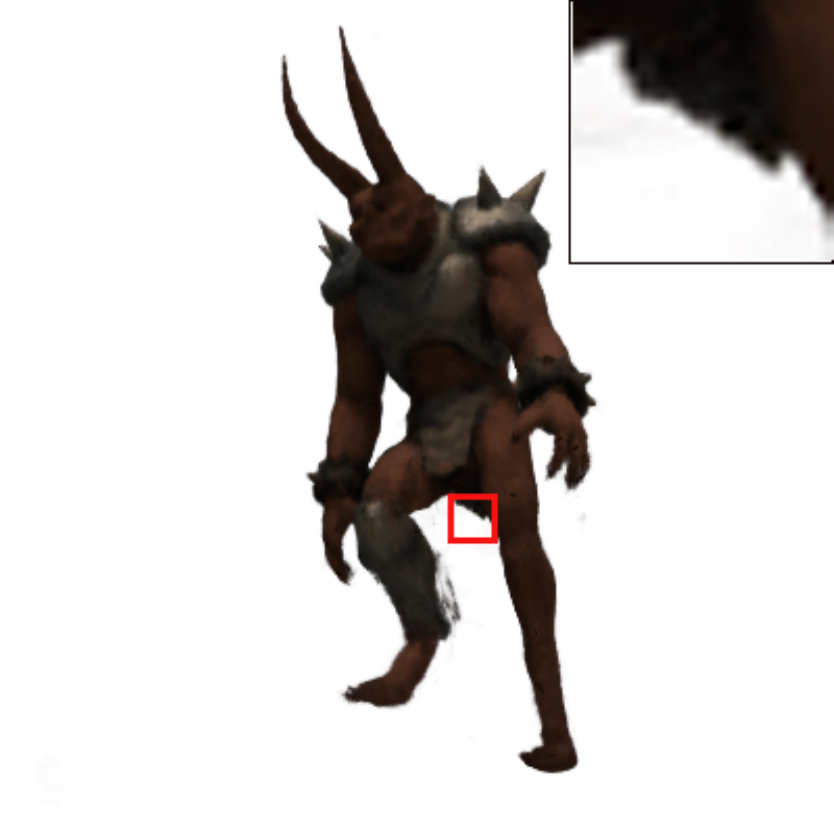}}
	\end{minipage}
	\begin{minipage}[t]{0.23\linewidth}
		\centerline{\includegraphics[width=2.76cm]{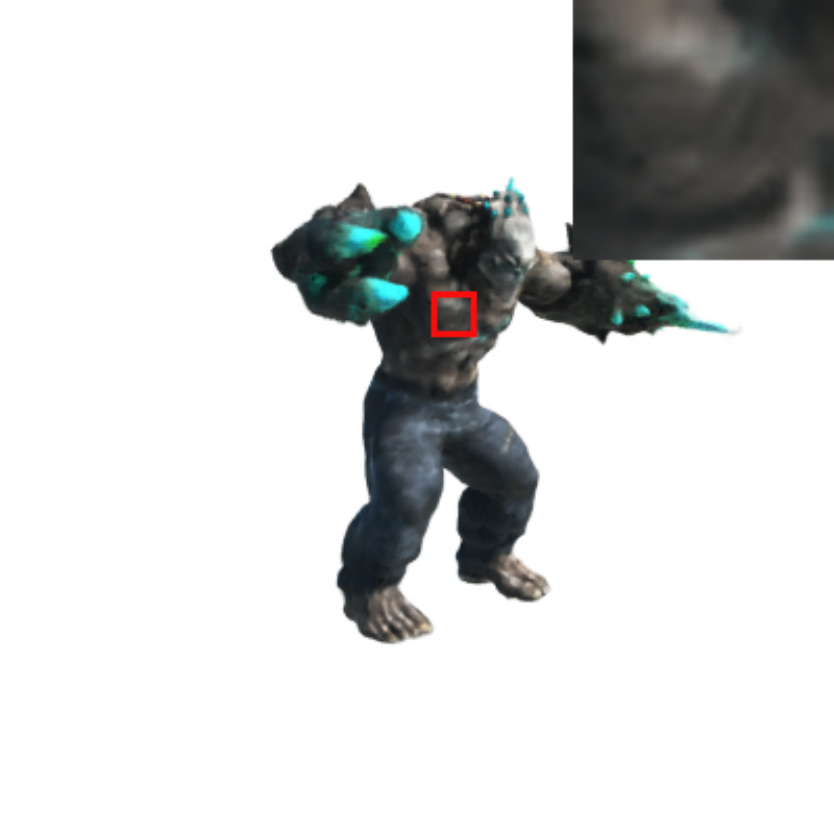}}
	\end{minipage}
	\hfill
	\begin{minipage}[t]{0.03\linewidth}
	    \rotatebox{90}{\quad D-NeRF\cite{pumarola2021d}}
	\end{minipage}
	\begin{minipage}[t]{0.23\linewidth}
		\centerline{\includegraphics[width=2.76cm]{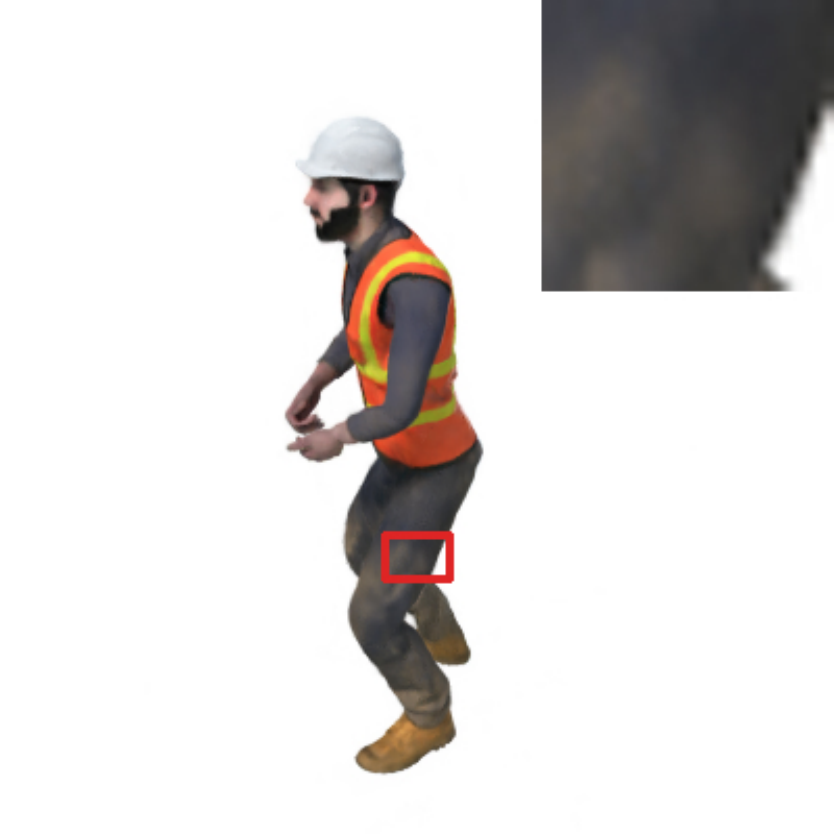}}
	\end{minipage}
	\begin{minipage}[t]{0.23\linewidth}
		\centerline{\includegraphics[width=2.76cm]{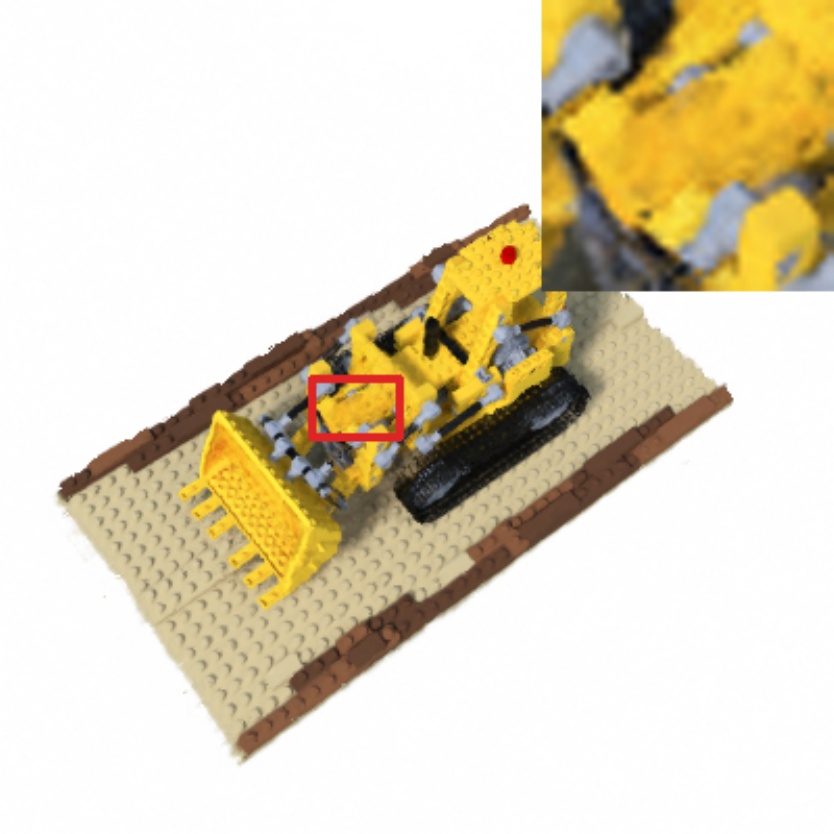}}
	\end{minipage}
	\begin{minipage}[t]{0.23\linewidth}
		\centerline{\includegraphics[width=2.76cm]{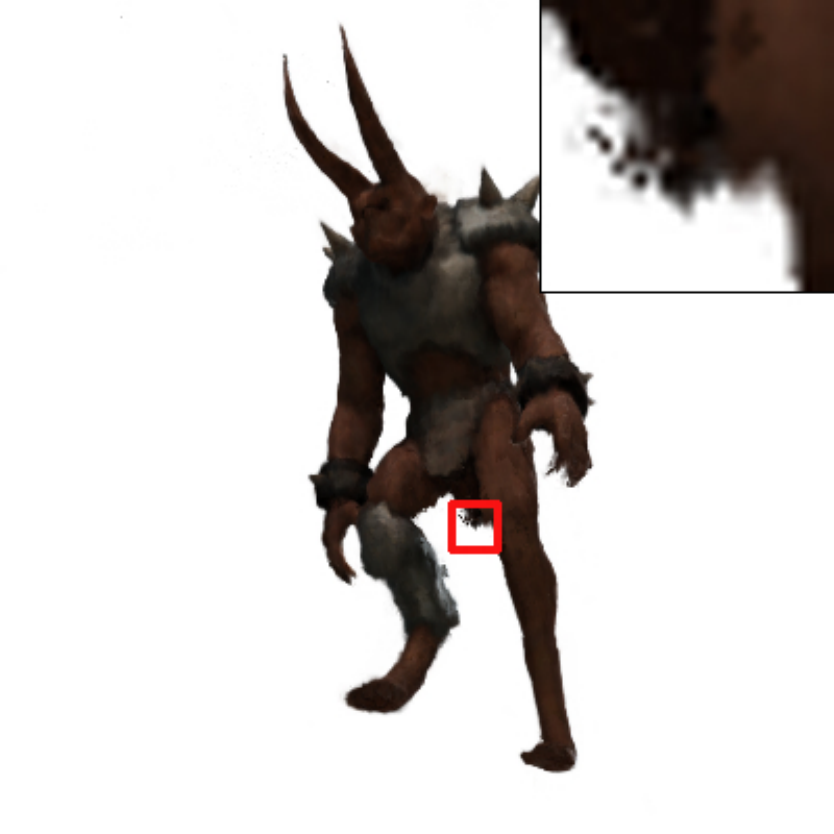}}
	\end{minipage}
	\begin{minipage}[t]{0.23\linewidth}
		\centerline{\includegraphics[width=2.76cm]{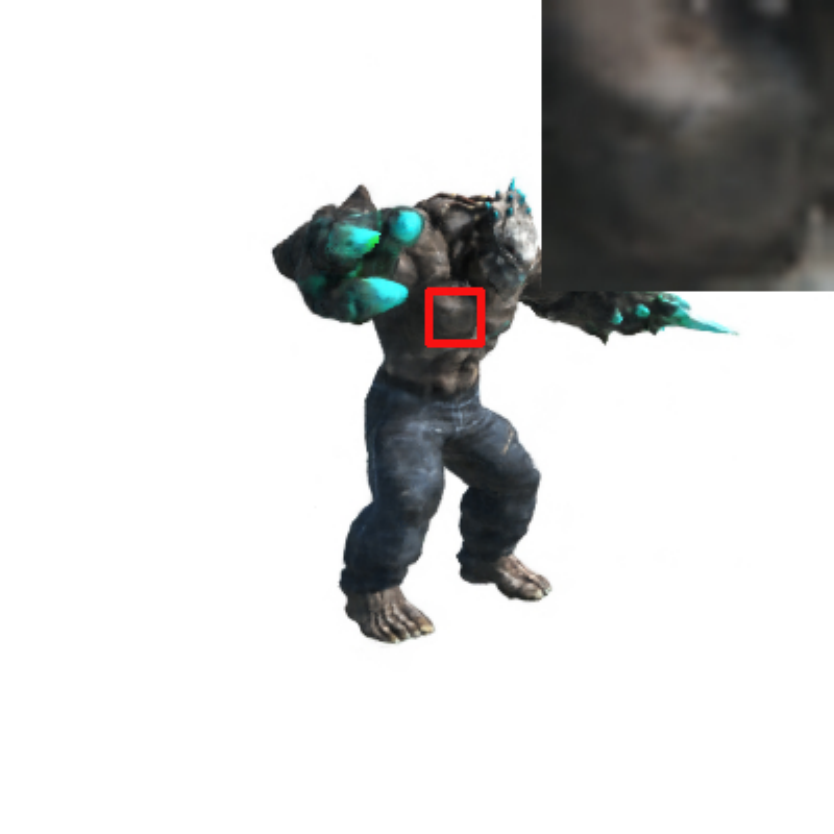}}
	\end{minipage}
	\begin{minipage}[t]{0.03\linewidth}
		\rotatebox{90}{\qquad Ours}
	\end{minipage}
	\begin{minipage}[t]{0.23\linewidth}
		\centerline{\includegraphics[width=2.76cm]{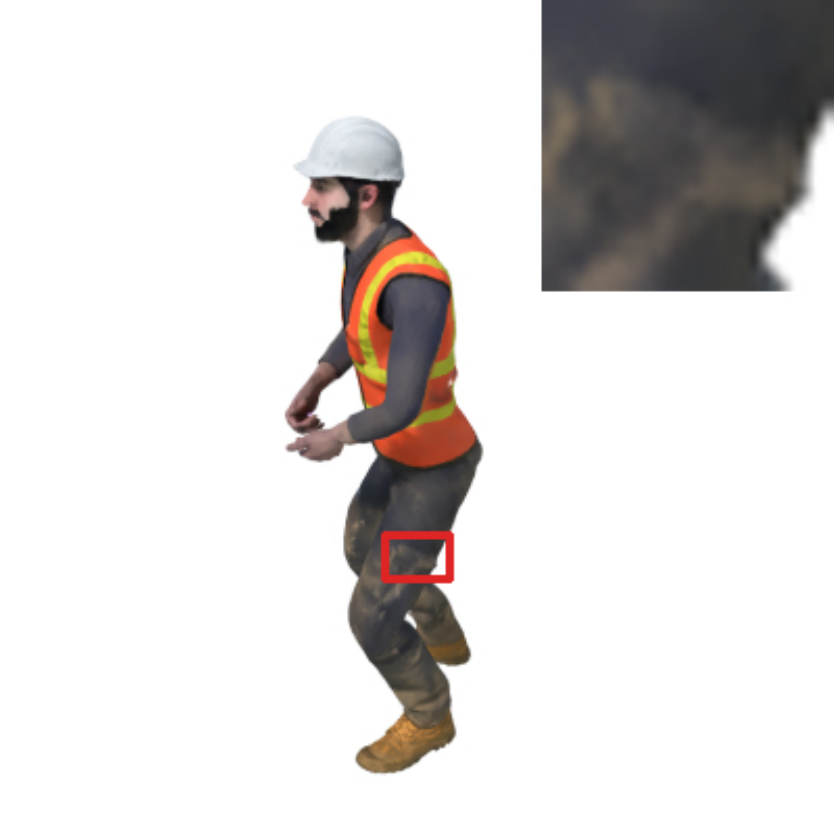}}
	\end{minipage}
	\begin{minipage}[t]{0.23\linewidth}
		\centerline{\includegraphics[width=2.76cm]{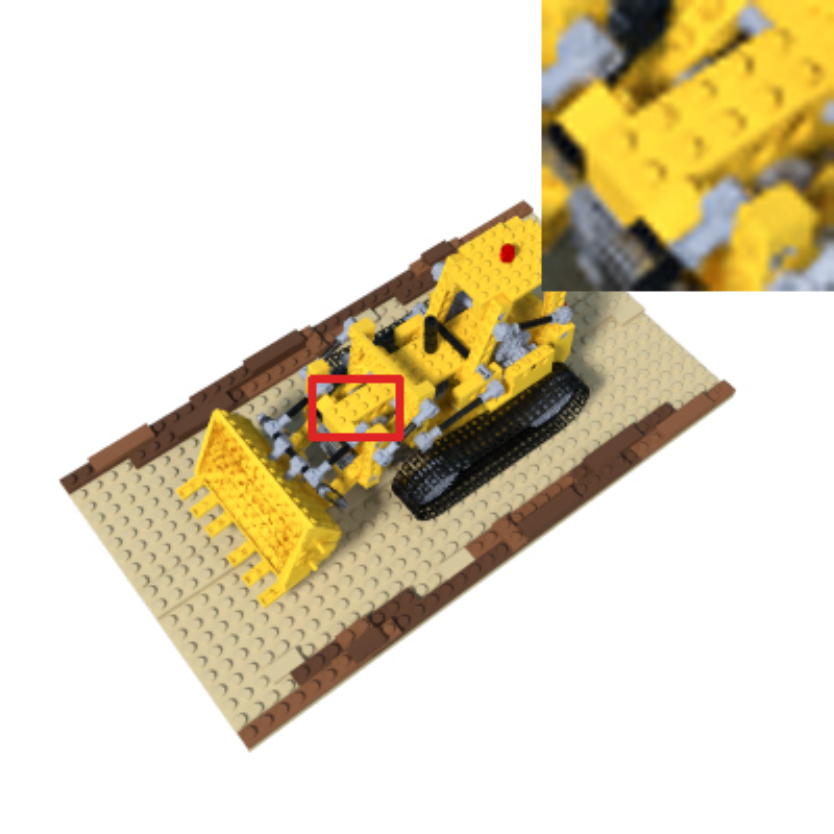}}
	\end{minipage}
	\begin{minipage}[t]{0.23\linewidth}
		\centerline{\includegraphics[width=2.76cm]{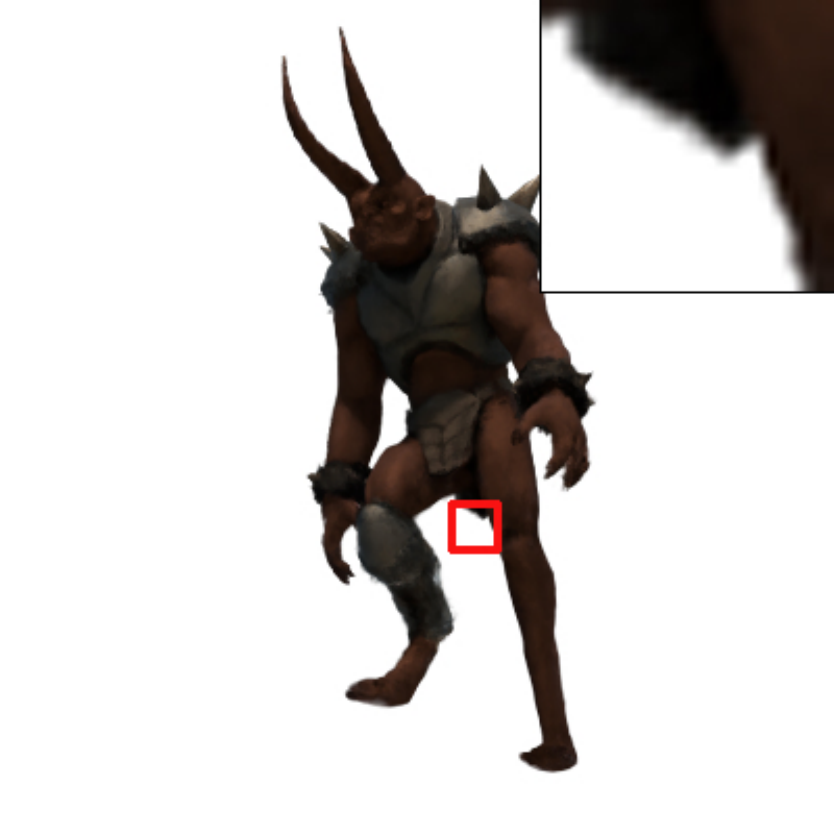}}
	\end{minipage}
	\begin{minipage}[t]{0.23\linewidth}
		\centerline{\includegraphics[width=2.76cm]{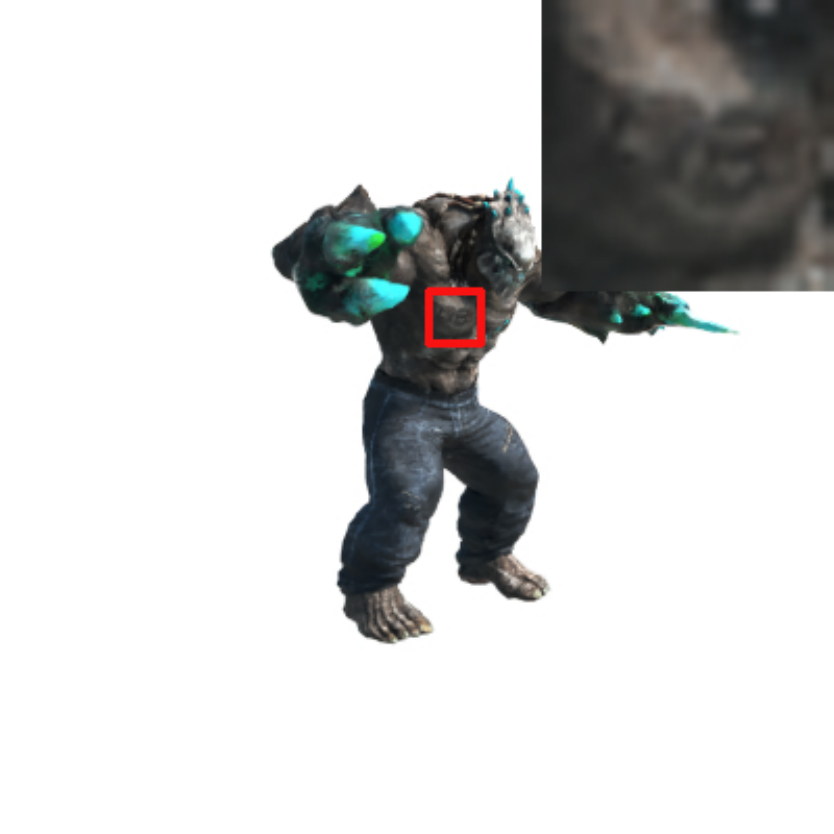}}
	\end{minipage}
	\begin{minipage}[t]{0.03\linewidth}
		\rotatebox{90}{\qquad GT}
	\end{minipage}
	\begin{minipage}[t]{0.23\linewidth}
		\centerline{\includegraphics[width=2.76cm]{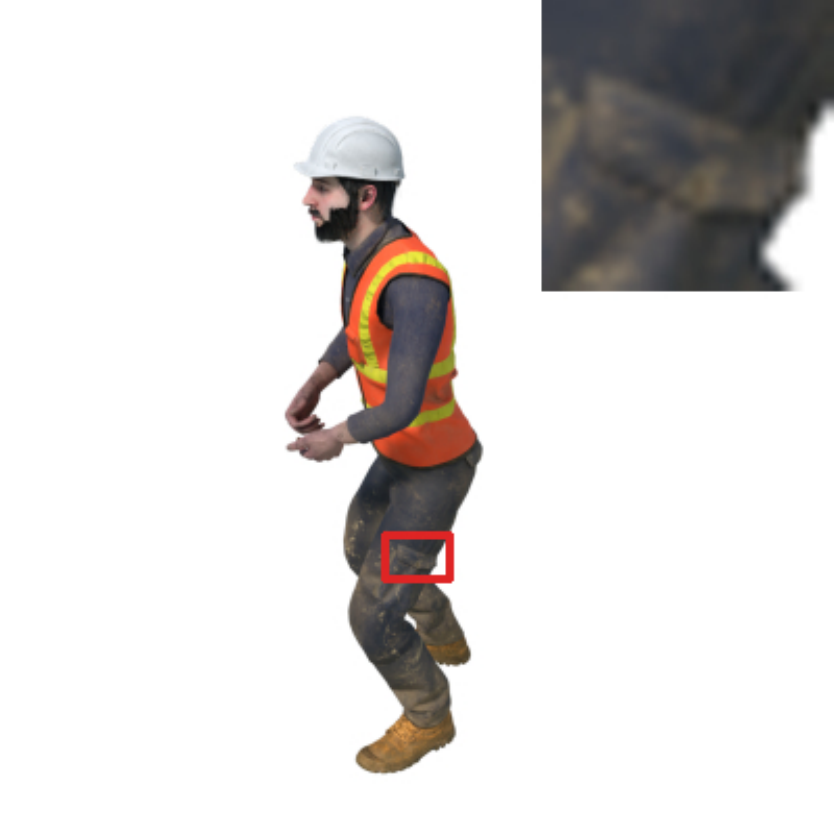}}
	\end{minipage}
	\begin{minipage}[t]{0.23\linewidth}
		\centerline{\includegraphics[width=2.76cm]{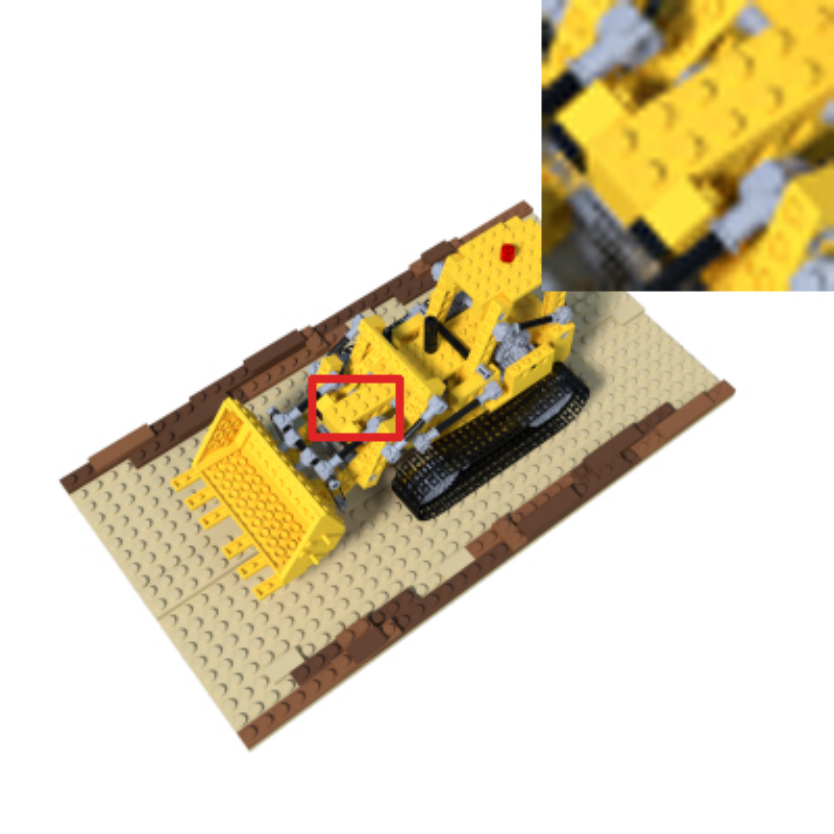}}
	\end{minipage}
	\begin{minipage}[t]{0.23\linewidth}
		\centerline{\includegraphics[width=2.76cm]{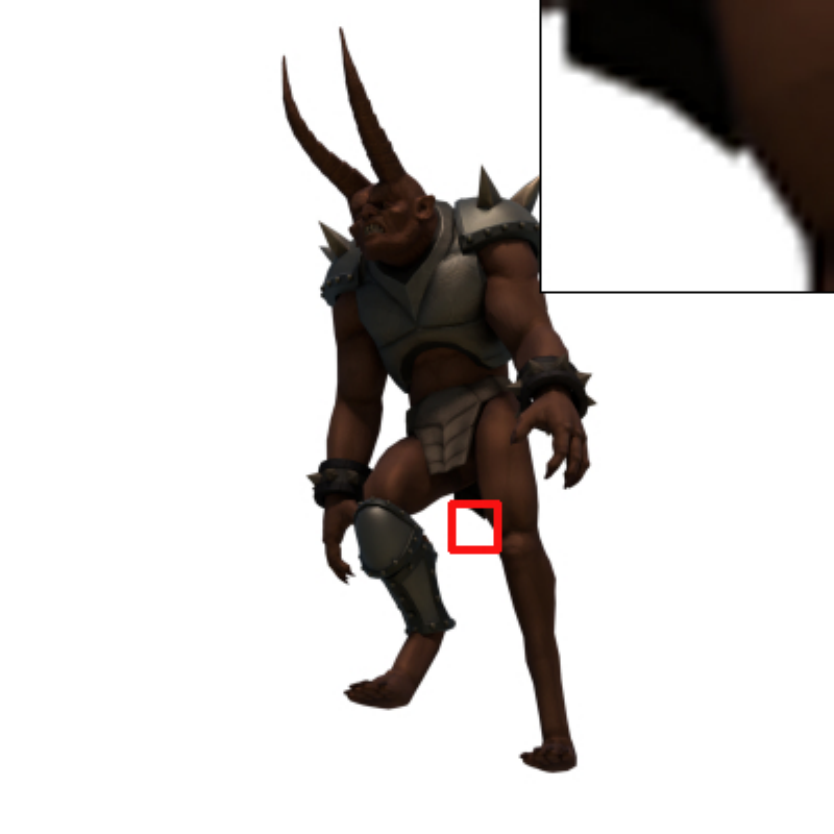}}
	\end{minipage}
	\begin{minipage}[t]{0.23\linewidth}
		\centerline{\includegraphics[width=2.76cm]{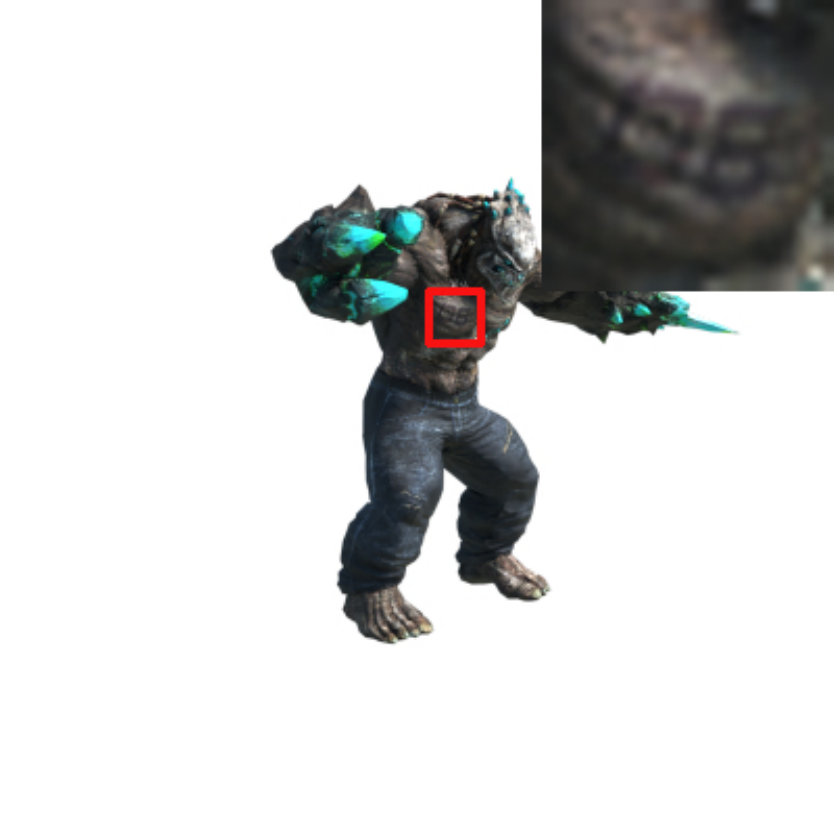}}
	\end{minipage}
	\begin{minipage}[t]{0.03\linewidth}
	    \rotatebox{90}{\quad TiNeuVox \cite{TiNeuVox}}
	\end{minipage}
	\begin{minipage}[t]{0.23\linewidth}
		\centerline{\includegraphics[width=2.76cm]{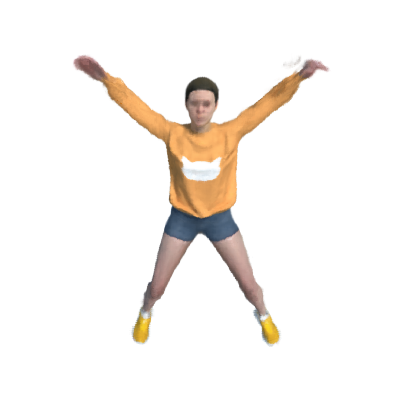}}
	\end{minipage}
	\begin{minipage}[t]{0.23\linewidth}
		\centerline{\includegraphics[width=2.76cm]{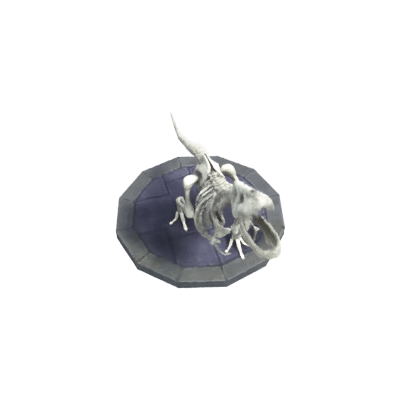}}
	\end{minipage}
	\begin{minipage}[t]{0.23\linewidth}
		\centerline{\includegraphics[width=2.76cm]{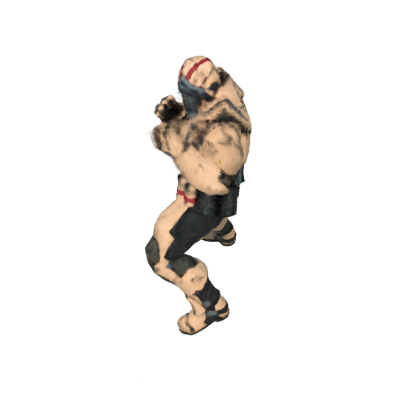}}
	\end{minipage}
	\begin{minipage}[t]{0.23\linewidth}
		\centerline{\includegraphics[width=2.76cm]{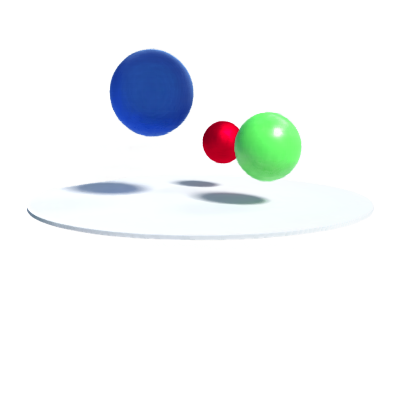}}
	\end{minipage}
	\begin{minipage}[t]{0.03\linewidth}
	    \rotatebox{90}{\quad D-NeRF\cite{pumarola2021d}}
	\end{minipage}
	\begin{minipage}[t]{0.23\linewidth}
		\centerline{\includegraphics[width=2.76cm]{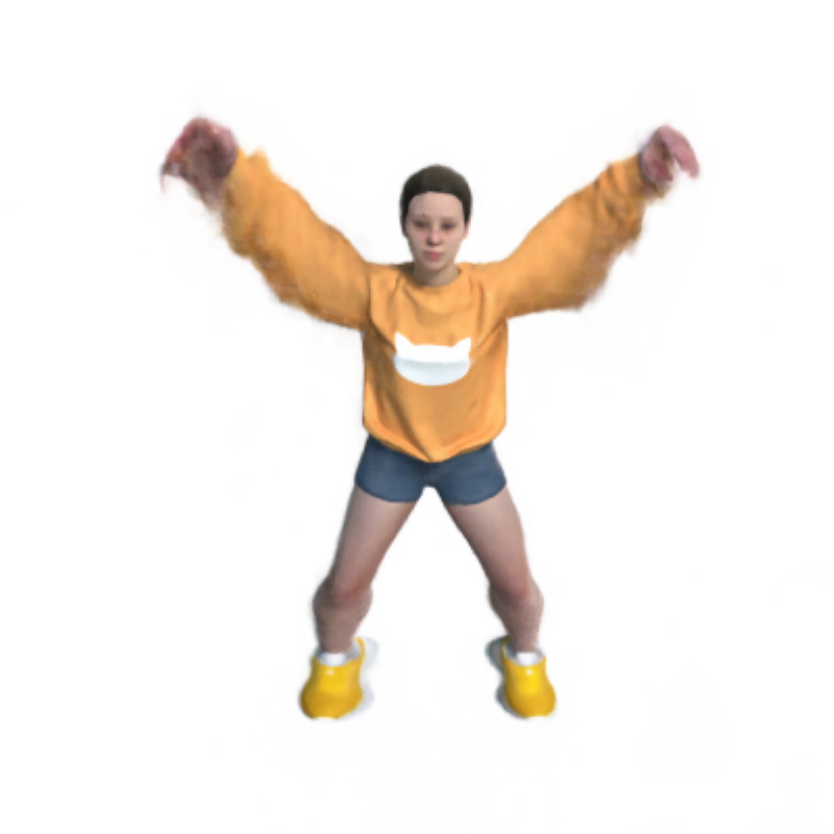}}
	\end{minipage}
	\begin{minipage}[t]{0.23\linewidth}
		\centerline{\includegraphics[width=2.76cm]{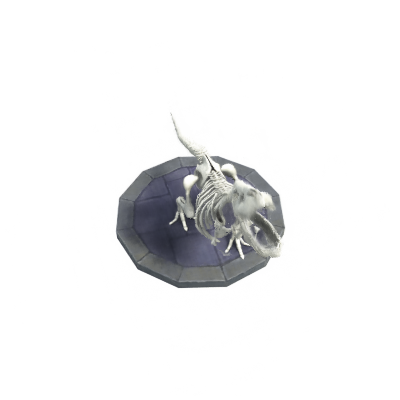}}
	\end{minipage}
	\begin{minipage}[t]{0.23\linewidth}
		\centerline{\includegraphics[width=2.76cm]{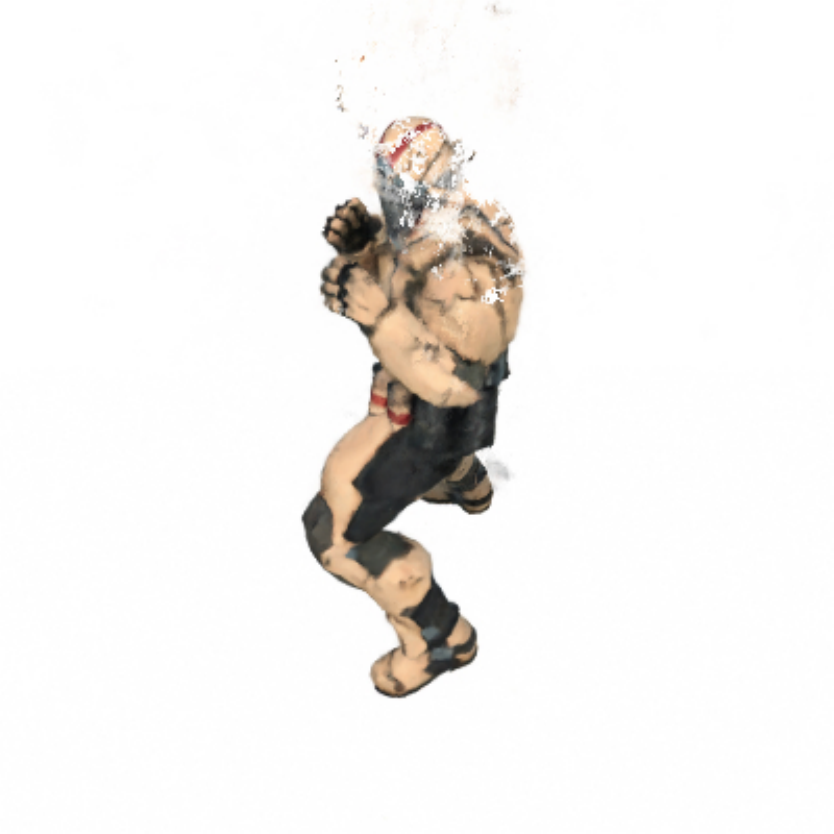}}
	\end{minipage}
	\begin{minipage}[t]{0.23\linewidth}
		\centerline{\includegraphics[width=2.76cm]{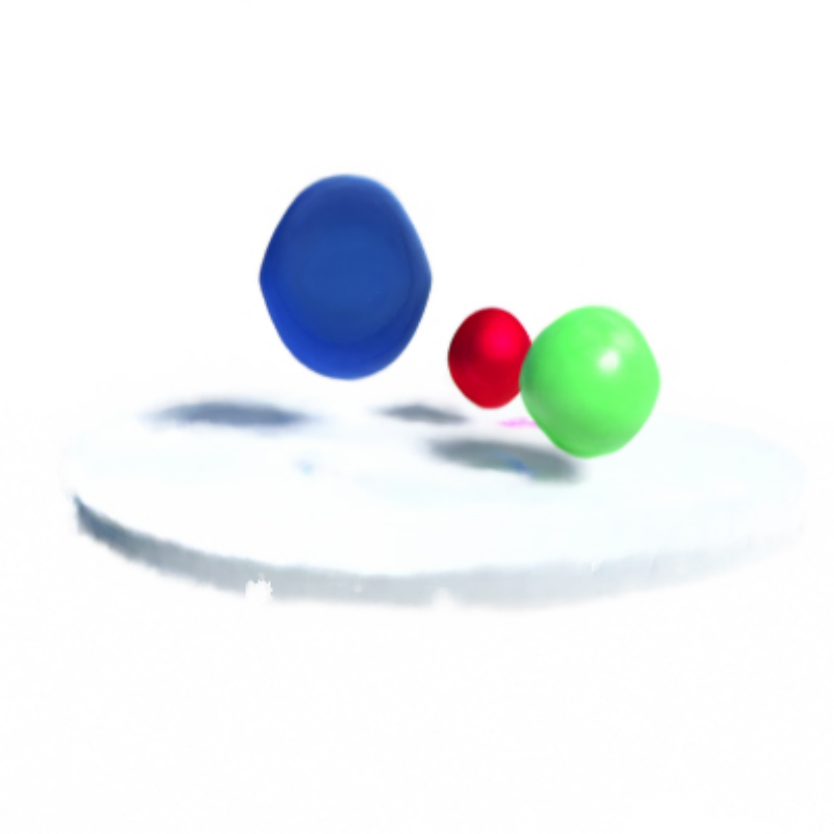}}
	\end{minipage}
	\begin{minipage}[t]{0.03\linewidth}
		\rotatebox{90}{\qquad Ours}
	\end{minipage}
	\begin{minipage}[t]{0.23\linewidth}
		\centerline{\includegraphics[width=2.76cm]{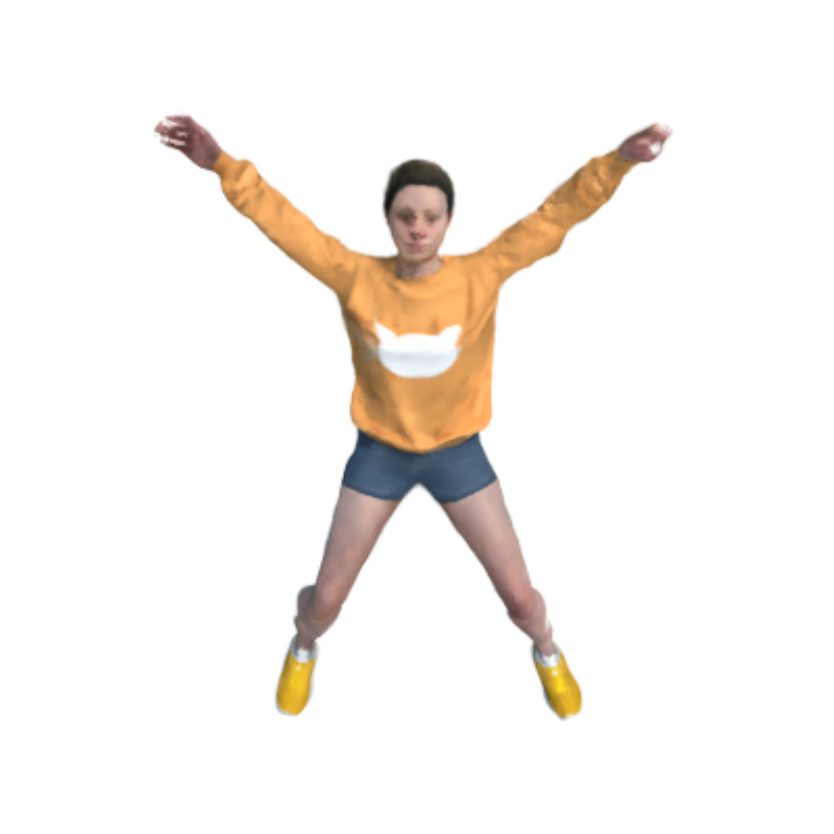}}
	\end{minipage}
	\begin{minipage}[t]{0.23\linewidth}
		\centerline{\includegraphics[width=2.76cm]{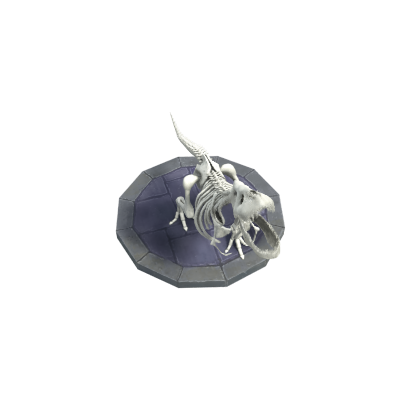}}
	\end{minipage}
	\begin{minipage}[t]{0.23\linewidth}
		\centerline{\includegraphics[width=2.76cm]{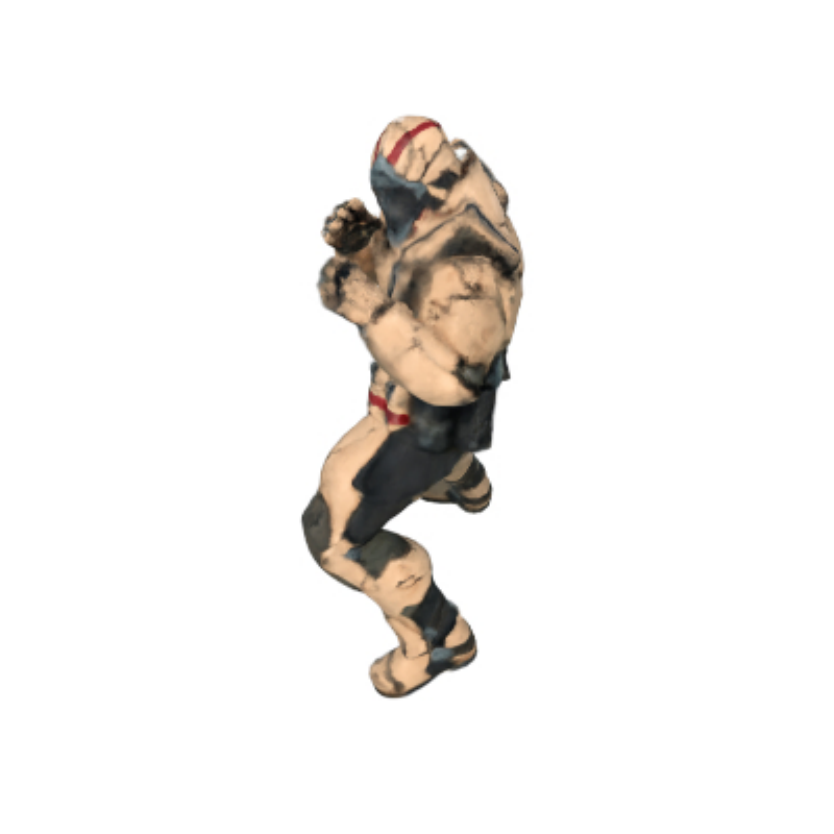}}
	\end{minipage}
	\begin{minipage}[t]{0.23\linewidth}
		\centerline{\includegraphics[width=2.76cm]{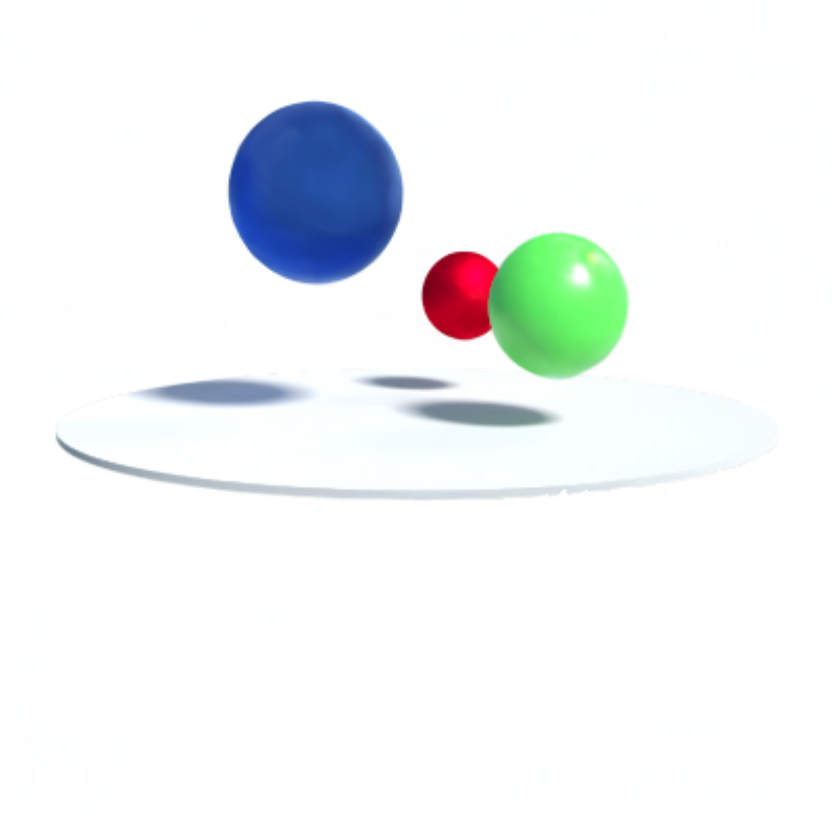}}
	\end{minipage}
	\begin{minipage}[t]{0.03\linewidth}
		\rotatebox{90}{\qquad GT}
	\end{minipage}
	\begin{minipage}[t]{0.23\linewidth}
		\centerline{\includegraphics[width=2.76cm]{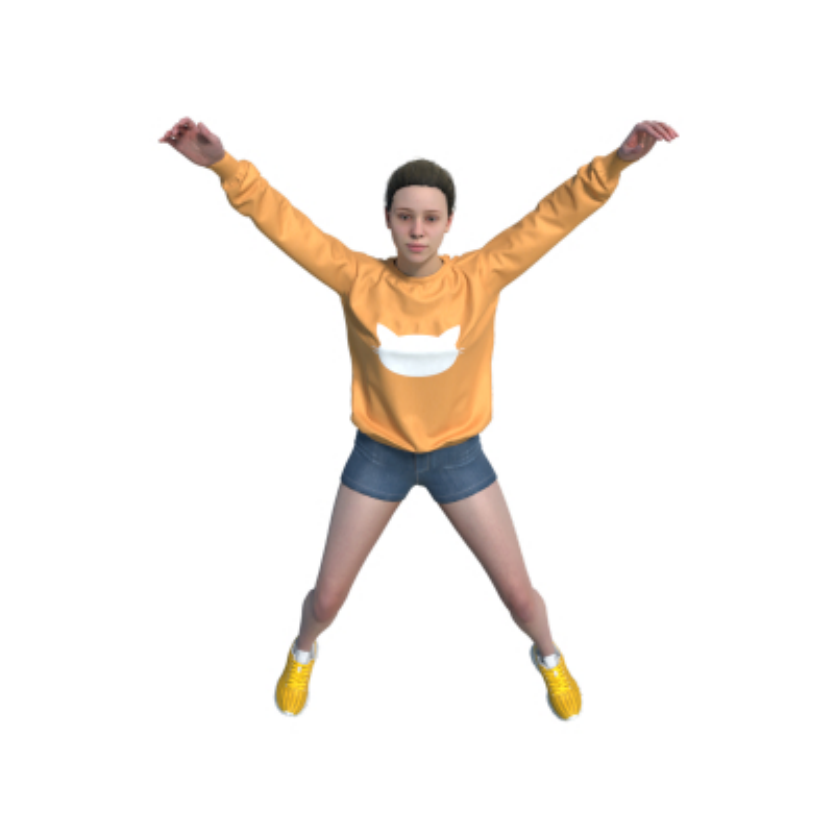}}
	\end{minipage}
	\begin{minipage}[t]{0.23\linewidth}
		\centerline{\includegraphics[width=2.76cm]{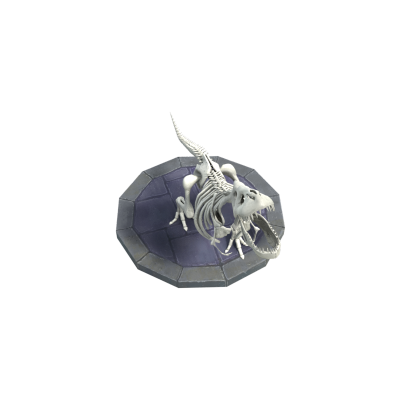}}
	\end{minipage}
	\begin{minipage}[t]{0.23\linewidth}
		\centerline{\includegraphics[width=2.76cm]{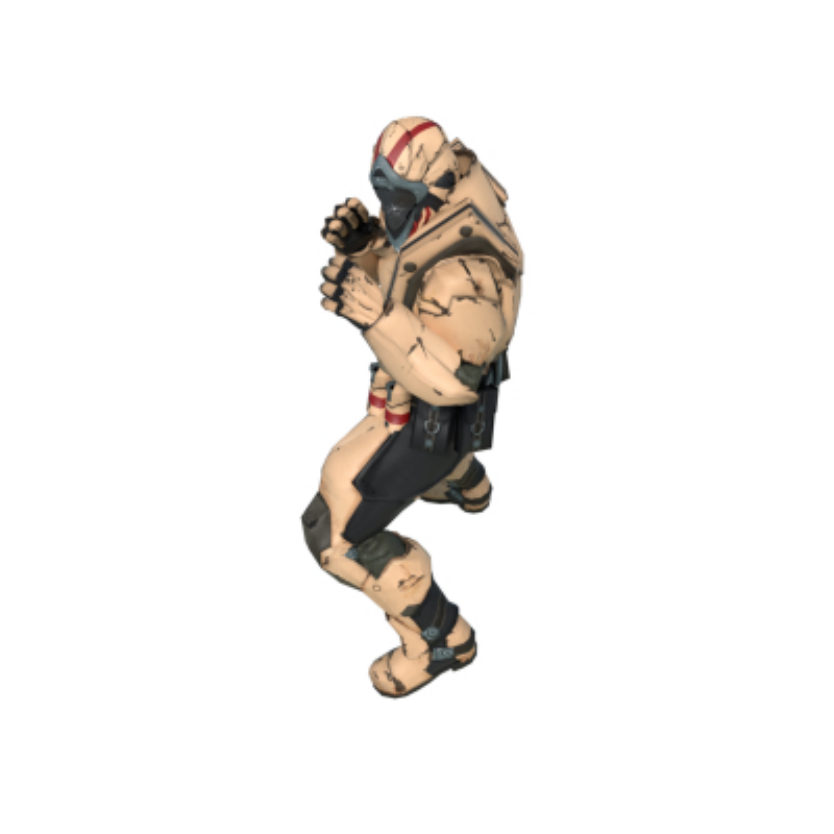}}
	\end{minipage}
	\begin{minipage}[t]{0.23\linewidth}
		\centerline{\includegraphics[width=2.76cm]{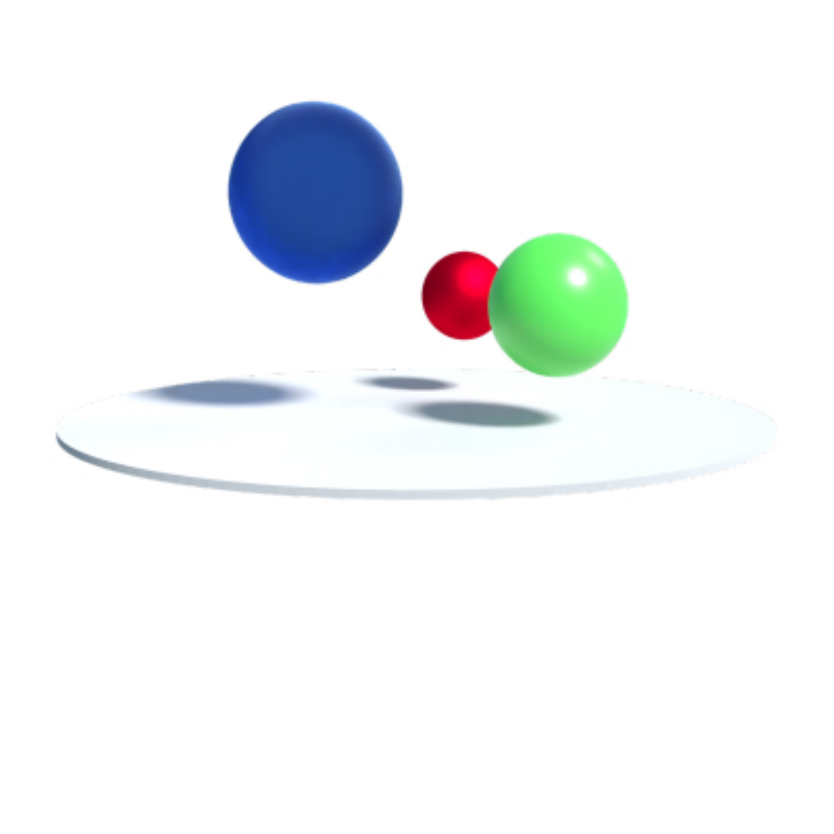}}
	\end{minipage}
	\centering
	\caption{Visual comparisons on the dynamic dataset. Please zoom in for better observation.}
	\label{benchmark_fig}
\end{figure*}

\begin{figure*}[t]
\setlength{\belowcaptionskip}{-0.3cm}
\centering
\includegraphics[width=1.0\textwidth]{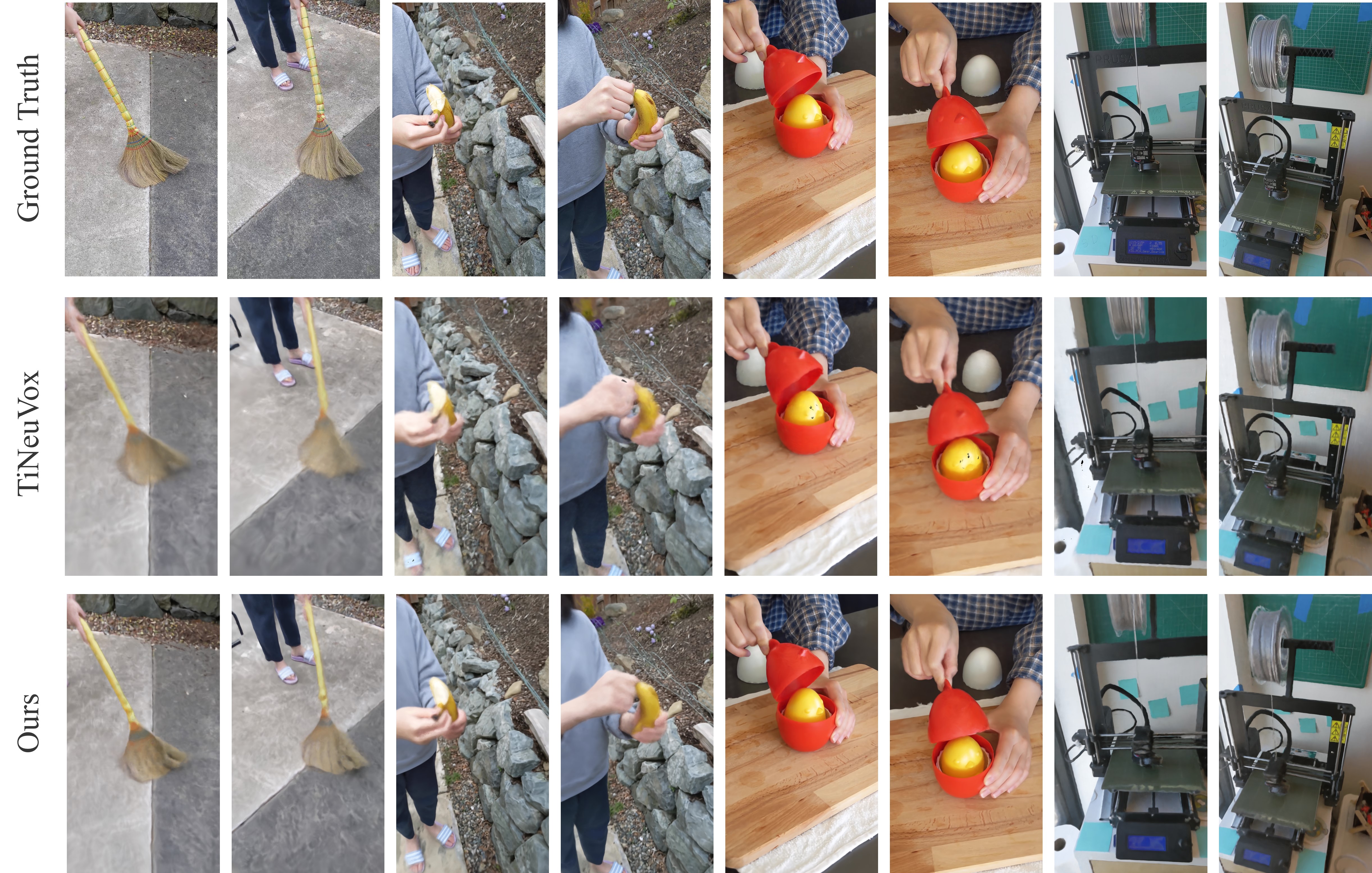}
\caption{Visual comparisons for ours and TiNeuVox \cite{TiNeuVox} on the dynamic real scenes \cite{park2021hypernerf}. Please zoom in for better observation.}
\label{fig:real scene}
\end{figure*}

\textbf{4D novel view synthesis}
The main motivation of our approach is to address the problems in the exited method D-NeRF \cite{pumarola2021d}. One problem is the unsatisfied performance due to the limited capacity in MLP-based networks, and the other problem is the high computational cost for both training and inference. Therefore, we strictly aligned the experiment setting with D-NeRF and compared the result with it. For the synthesis scenes \cite{pumarola2021d}, our proposed method is much better than the D-NeRF both quantitatively (Table \ref{t:benchmark_summarize}) and qualitatively (Figure \ref{benchmark_fig}). Besides, comparing the concurrent works, NDVG \cite{NDVG} and TiNeuVox \cite{TiNeuVox}, our method is also better than theirs. For the real dynamic scenes \cite{park2021hypernerf}, we also achieves competitive results as recorded in Table \ref{t:real_benchmark_summarize} and Figure \ref{fig:real scene}. The disadvantage of our method is that we need a longer training time, which is because we do not adapt the canonical manner and rely on the TV loss to pass through the temporal information between different frames, so we need more iterations to train the network. Though we have a longer training time, our inference time is lesser than TiNeuVox \cite{TiNeuVox} since we do not need to use an MLP-based sub-network to predict the position offset of each moment. The detailed numerical results are presented in the supplemental material.

\begin{figure*}[t]
\setlength{\belowcaptionskip}{0cm}
	\centering
	\begin{minipage}[t]{0.03\linewidth}
	    \rotatebox{90}{\qquad BL}
	\end{minipage}
	\begin{minipage}[t]{0.155\linewidth}
		\centerline{\includegraphics[width=2.2cm]{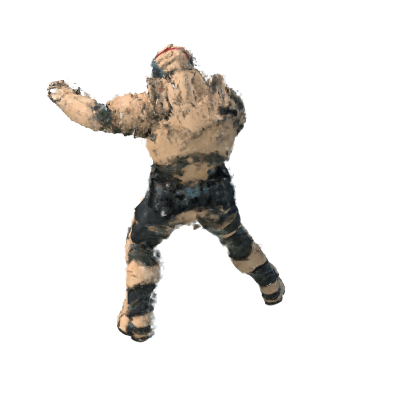}}
	\end{minipage}
	\begin{minipage}[t]{0.155\linewidth}
		\centerline{\includegraphics[width=2.2cm]{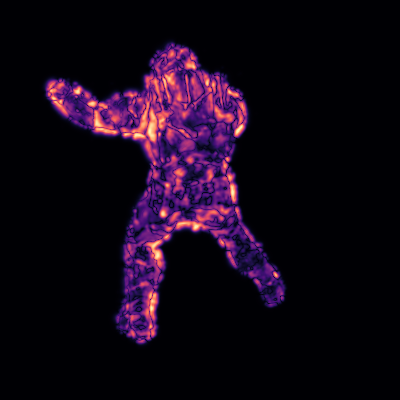}}
	\end{minipage}
	\begin{minipage}[t]{0.155\linewidth}
		\centerline{\includegraphics[width=2.2cm]{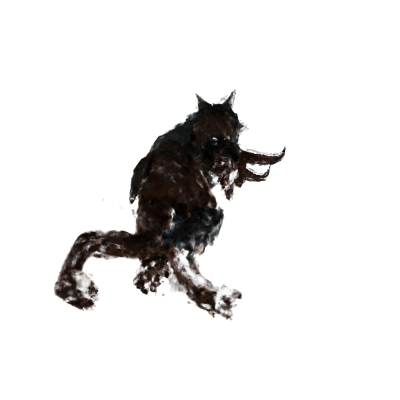}}
	\end{minipage}
	\begin{minipage}[t]{0.155\linewidth}
		\centerline{\includegraphics[width=2.2cm]{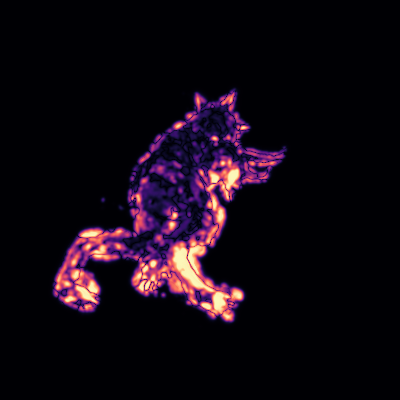}}
	\end{minipage}
	\begin{minipage}[t]{0.155\linewidth}
		\centerline{\includegraphics[width=2.2cm]{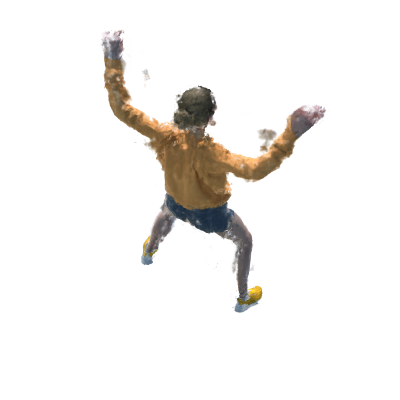}}
	\end{minipage}
	\begin{minipage}[t]{0.155\linewidth}
		\centerline{\includegraphics[width=2.2cm]{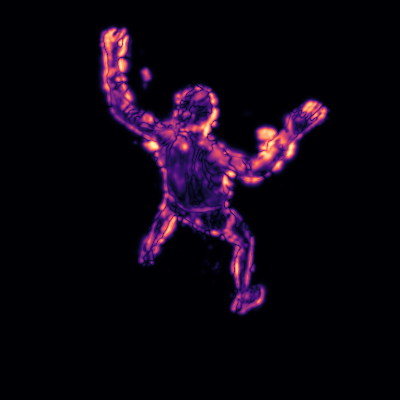}}
	\end{minipage}
	\hfill
	\begin{minipage}[t]{0.03\linewidth}
	    \rotatebox{90}{\qquad BL+D}
	\end{minipage}
	\begin{minipage}[t]{0.155\linewidth}
		\centerline{\includegraphics[width=2.2cm]{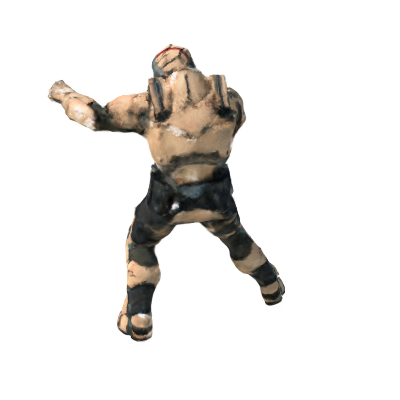}}
	\end{minipage}
	\begin{minipage}[t]{0.155\linewidth}
		\centerline{\includegraphics[width=2.2cm]{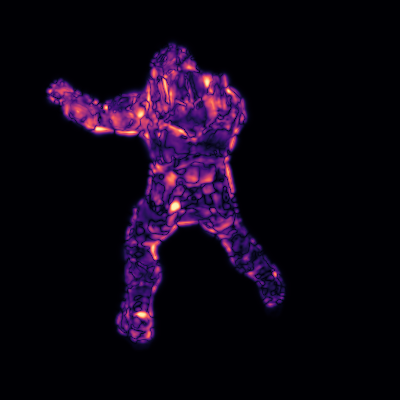}}
	\end{minipage}
	\begin{minipage}[t]{0.155\linewidth}
		\centerline{\includegraphics[width=2.2cm]{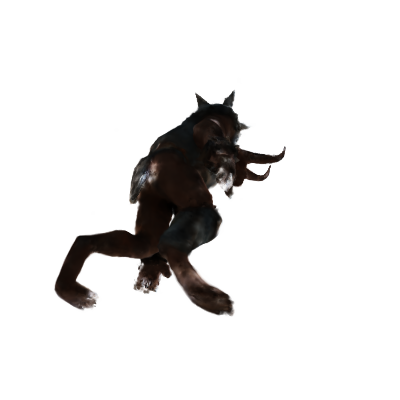}}
	\end{minipage}
	\begin{minipage}[t]{0.155\linewidth}
		\centerline{\includegraphics[width=2.2cm]{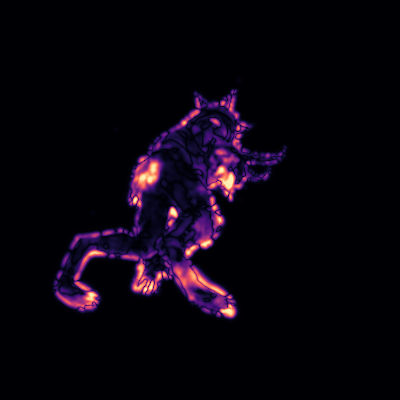}}
	\end{minipage}
	\begin{minipage}[t]{0.155\linewidth}
		\centerline{\includegraphics[width=2.2cm]{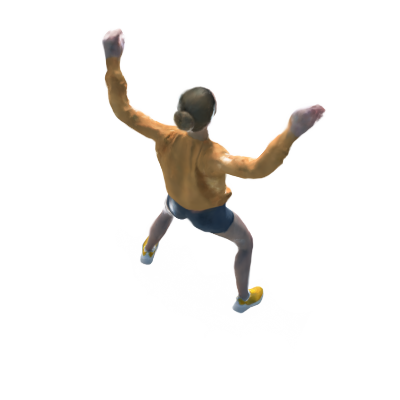}}
	\end{minipage}
	\begin{minipage}[t]{0.155\linewidth}
		\centerline{\includegraphics[width=2.2cm]{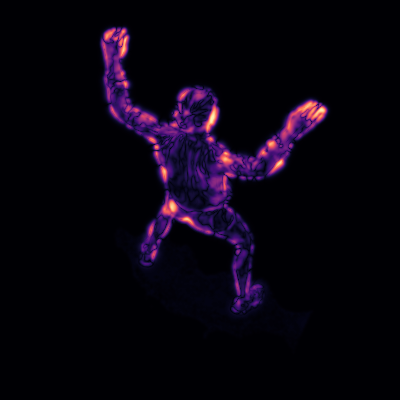}}
	\end{minipage}
	\hfill
	\begin{minipage}[t]{0.03\linewidth}
	    \rotatebox{90}{\quad BL+D+P}
	\end{minipage}
	\begin{minipage}[t]{0.155\linewidth}
		\centerline{\includegraphics[width=2.2cm]{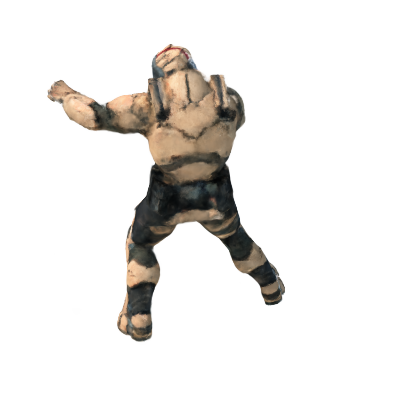}}
	\end{minipage}
	\begin{minipage}[t]{0.155\linewidth}
		\centerline{\includegraphics[width=2.2cm]{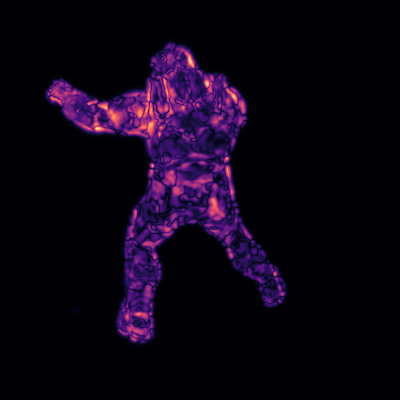}}
	\end{minipage}
	\begin{minipage}[t]{0.155\linewidth}
		\centerline{\includegraphics[width=2.2cm]{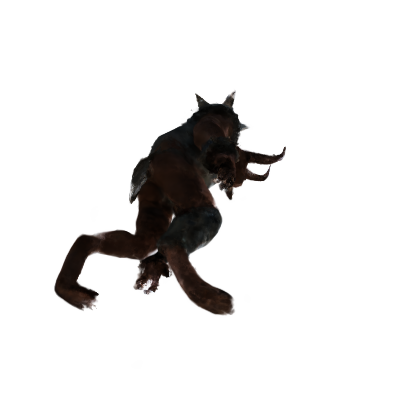}}
	\end{minipage}
	\begin{minipage}[t]{0.155\linewidth}
		\centerline{\includegraphics[width=2.2cm]{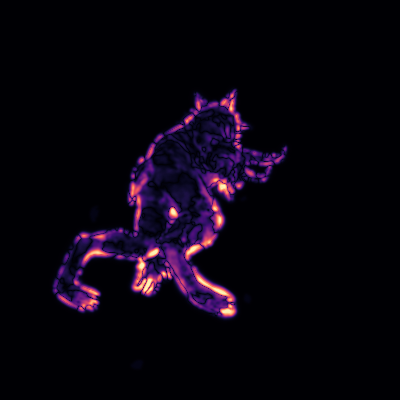}}
	\end{minipage}
	\begin{minipage}[t]{0.155\linewidth}
		\centerline{\includegraphics[width=2.2cm]{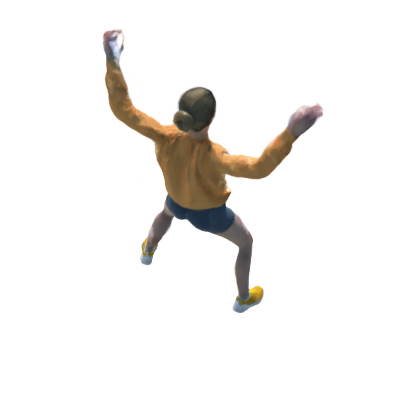}}
	\end{minipage}
	\begin{minipage}[t]{0.155\linewidth}
		\centerline{\includegraphics[width=2.2cm]{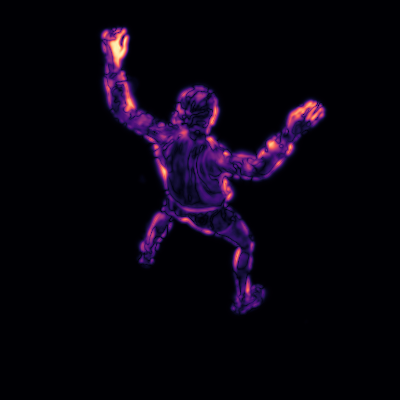}}
	\end{minipage}
	\hfill
	\begin{minipage}[t]{0.03\linewidth}
	    \rotatebox{90}{\quad BL+D+C}
	\end{minipage}
	\begin{minipage}[t]{0.155\linewidth}
		\centerline{\includegraphics[width=2.2cm]{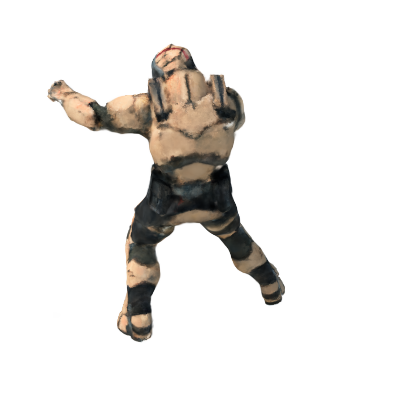}}
	\end{minipage}
	\begin{minipage}[t]{0.155\linewidth}
		\centerline{\includegraphics[width=2.2cm]{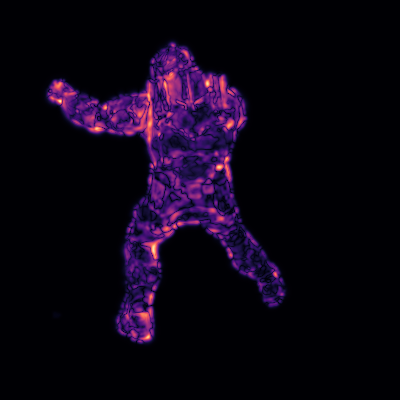}}
	\end{minipage}
	\begin{minipage}[t]{0.155\linewidth}
		\centerline{\includegraphics[width=2.2cm]{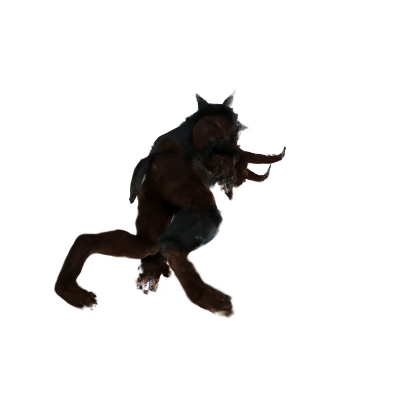}}
	\end{minipage}
	\begin{minipage}[t]{0.155\linewidth}
		\centerline{\includegraphics[width=2.2cm]{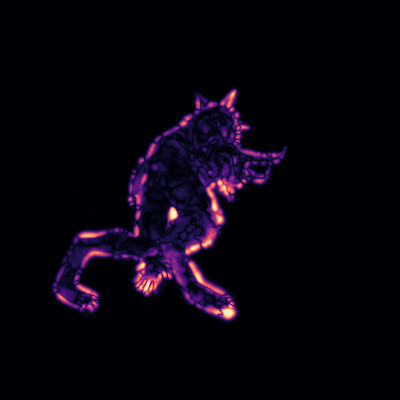}}
	\end{minipage}
	\begin{minipage}[t]{0.155\linewidth}
		\centerline{\includegraphics[width=2.2cm]{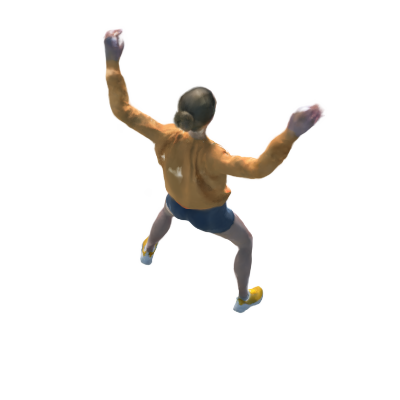}}
	\end{minipage}
	\begin{minipage}[t]{0.155\linewidth}
		\centerline{\includegraphics[width=2.2cm]{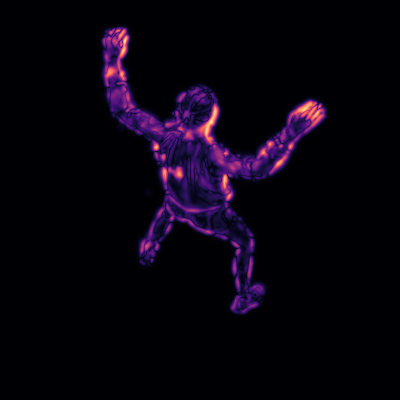}}
	\end{minipage}
	\hfill
	\begin{minipage}[t]{0.03\linewidth}
	    \rotatebox{90}{BL+D+C+L}
	\end{minipage}
	\begin{minipage}[t]{0.155\linewidth}
		\centerline{\includegraphics[width=2.2cm]{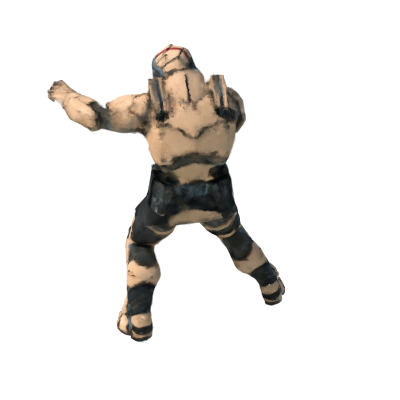}}
		\centerline{Hook}
	\end{minipage}
	\begin{minipage}[t]{0.155\linewidth}
		\centerline{\includegraphics[width=2.2cm]{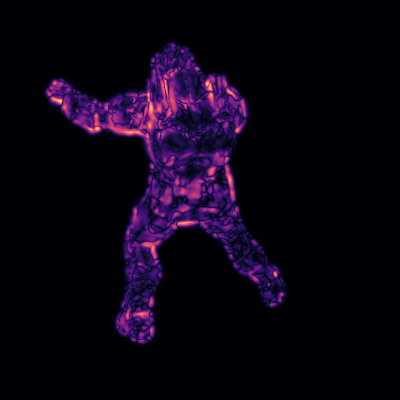}}
	\end{minipage}
	\begin{minipage}[t]{0.155\linewidth}
		\centerline{\includegraphics[width=2.2cm]{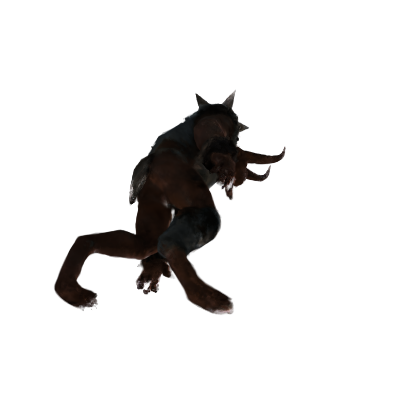}}
		\centerline{Hell Warrior}
	\end{minipage}
	\begin{minipage}[t]{0.155\linewidth}
		\centerline{\includegraphics[width=2.2cm]{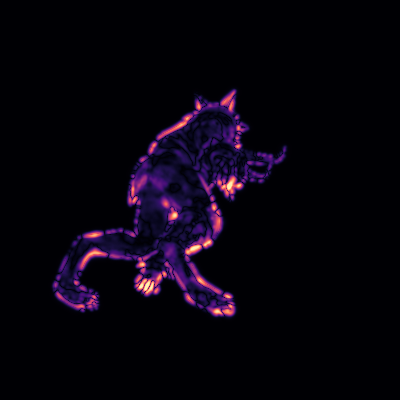}}
	\end{minipage}
	\begin{minipage}[t]{0.155\linewidth}
		\centerline{\includegraphics[width=2.2cm]{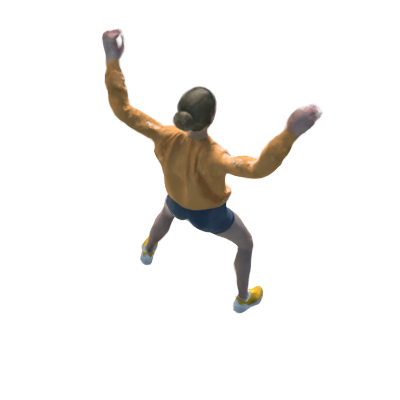}}
		\centerline{Jumping Jacks}
	\end{minipage}
	\begin{minipage}[t]{0.155\linewidth}
		\centerline{\includegraphics[width=2.2cm]{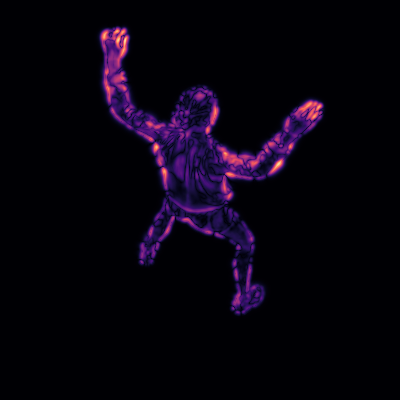}}
	\end{minipage}
	\hfill
	\centering
	\caption{FLIP error map \cite{flip} visualization on the ablation studies about neural network architecture. The brighter color means the larger error. Please zoom in for better observation.}\label{flip on 4d}
\end{figure*}

\begin{table*}[t]
\renewcommand\arraystretch{1.25}
\setlength{\belowcaptionskip}{0.1cm}
\centering
\caption{Ablation study of the V4D architecture. The definition of each abbreviation is SV (single voxel that combines the density volume and texture volume ), SF (single radiance field that combines the density field and texture field), BL (our proposed baseline model), D (using the decay total variation loss), P (using the positional encoding in the sampled voxel feature), C (using the conditional positional encoding in the sampled voxel feature), L (using the LUTs refinement module). The partial visualization result is in Figure \ref{flip on 4d}. The number with bold typeface means the best result.}
\scalebox{0.68}{
\begin{tabular}{lcccccccccccc}

\hline
\multirow{2}{*}{Method} & \multicolumn{3}{c}{{Hell Warrior}}  &  \multicolumn{3}{c}{{Mutant}}  & \multicolumn{3}{c}{{Hook}} & \multicolumn{3}{c}{{Bouncing Balls}}  \\

  & PSNR $\uparrow$ & SSIM $\uparrow$ & LPIPS $\downarrow$ & PSNR $\uparrow$ & SSIM $\uparrow$ & LPIPS $\downarrow$  & PSNR $\uparrow$ & SSIM $\uparrow$ & LPIPS $\downarrow$  & PSNR $\uparrow$ & SSIM $\uparrow$ & LPIPS $\downarrow$

\tabularnewline
\hline

SV & 25.81 & 0.95 & 0.06  &  35.82  & 0.99  & 0.01 &  29.93 & 0.96 & 0.04 & 41.90 & 0.99 & 0.02

\tabularnewline
SF & 25.93 & 0.95 & 0.07  & 36.08  & 0.99  & 0.01 & 30.14 & 0.97 & 0.04 & 30.10 & 0.97 & 0.07

\tabularnewline
\hline

BL & 23.12 & 0.93  & 0.09  & 34.64   &  0.98 & 0.02 & 27.85 & 0.95 & 0.06  & 37.60 & 0.99 & 0.03 

\tabularnewline

+ D & 26.04 & 0.95  & 0.06   & 36.03   & 0.99  & 0.02 & 30.27 & 0.96 & 0.04  & 40.28  & 0.99  & 0.03  
\tabularnewline

+ (D,P) & 26.57 & 0.96 & 0.06   &  36.18  & 0.99  & 0.02  & 30.43 & 0.96 &  0.04 & 40.96 & 0.99 &  0.02

\tabularnewline

+ (D,C) &  26.79 & 0.96 &  0.06 &  36.12  & 0.99  & 0.01 &   30.85 & 0.97  & 0.04 & 41.62 & 0.99 & 0.02


\tabularnewline
\hline

+ (D,C,L)  &  \textbf{27.03} & \textbf{0.96} & \textbf{0.05}  &  \textbf{36.27} & \textbf{0.99} & \textbf{0.01} & \textbf{31.04}  & \textbf{0.97} & \textbf{0.03} & \textbf{42.67} & \textbf{0.99} & \textbf{0.02}

\tabularnewline
\hline
\multirow{2}{*}{Method} & \multicolumn{3}{c}{{Lego}}  &  \multicolumn{3}{c}{{T-Rex}}  & \multicolumn{3}{c}{{Stand Up}} & \multicolumn{3}{c}{{Jumping Jacks}}  \\ 

 & PSNR $\uparrow$ & SSIM $\uparrow$ & LPIPS $\downarrow$ & PSNR $\uparrow$ & SSIM $\uparrow$ & LPIPS $\downarrow$  & PSNR $\uparrow$ & SSIM $\uparrow$ & LPIPS $\downarrow$  & PSNR $\uparrow$ & SSIM $\uparrow$ & LPIPS $\downarrow$

\tabularnewline
\hline
SV & \textbf{25.63} & \textbf{0.95} & \textbf{0.04}  &  34.18  & 0.98  & 0.02 & 36.96  & 0.99 & 0.01 & 35.31 & 0.99 & 0.02

\tabularnewline
SF & 25.17 & 0.94 & 0.04  & 32.60  & 0.98  & 0.02 & 37.15 & 0.99 & 0.01& \textbf{35.44} & \textbf{0.99} & \textbf{0.02}

\tabularnewline
\hline

BL & 25.17 & 0.94 & 0.05  & 32.74   & 0.98  & 0.03  & 33.26 & 0.98 & 0.02  & 30.77 & 0.97 & 0.04 

\tabularnewline

+ D & 25.57 & 0.95 & 0.04   &  33.91  & 0.98  & 0.02 & 36.66 & 0.99 & 0.01  & 34.81 & 0.98 & 0.02 
\tabularnewline

+ (D,P) & 25.62 & 0.95 & 0.04   & \textbf{34.56}   & \textbf{0.99}  & \textbf{0.02} & 36.91 & 0.99 & 0.01  & 34.97 & 0.98  & 0.02  

\tabularnewline

+ (D,C) & 25.61 & 0.95 & 0.04 &  34.48  & 0.99  & 0.02 & 37.11 & 0.99 & 0.01  & 35.18 & 0.99 & 0.02  

\tabularnewline
\hline
+ (D,C,L)  & 25.62 & 0.95 & 0.04  & 34.53  & 0.99 & 0.02  & \textbf{37.20}  & \textbf{0.99} & \textbf{0.01} & 35.36 & 0.99 & 0.02

\tabularnewline
\hline

\end{tabular}
}

\label{t:v4d ablation}
\end{table*}

\begin{table*}[t]
\renewcommand\arraystretch{1.25}
\setlength{\belowcaptionskip}{0.1cm}
\centering
\caption{The hyperparameters ablation study of the LUTs refinement module. We select the DVGO \cite{sun2021direct} as the baseline method and evaluate 8 models in Synthetic-NeRF dataset \cite{nerf}. The definition of each abbreviation is \textbf{w/o weight} (without using the weight estimated by the pseudo-surface), \textbf{iter/iters} (the times of the recurrent iteration), \textbf{basic/basics} (the number of the basic LUT). Based on the overall evaluation of the three metrics, the number with bold typeface means the best and the number with the underline is the second.}
\setlength{\tabcolsep}{1.0mm}{}
\scalebox{0.75}{
\begin{tabular}{lcccccccccccc}

\hline
\multirow{2}{*}{Method} & \multicolumn{3}{c}{{Chair}}  &  \multicolumn{3}{c}{{Drums}}  & \multicolumn{3}{c}{{Ficus}} & \multicolumn{3}{c}{{Hotdogs}} \\

 & PSNR $\uparrow$ & SSIM $\uparrow$ & LPIPS $\downarrow$ & PSNR $\uparrow$ & SSIM $\uparrow$ & LPIPS $\downarrow$  & PSNR $\uparrow$ & SSIM $\uparrow$ & LPIPS $\downarrow$  & PSNR $\uparrow$ & SSIM $\uparrow$ & LPIPS $\downarrow$

\tabularnewline
\hline

DVGO \cite{sun2021direct}  & 34.455  &  0.979 & 0.022 & 25.514  & 0.930 & 0.074 & 32.914 & 0.978 & 0.024  & 36.997  &  0.981 & 0.031

\tabularnewline
\hline
+ LUT \\ (w/o weight)  & 34.581  & 0.980 & 0.024 & 25.367  & 0.930 & 0.078  & 33.640  & 0.981 & 0.023  & 37.052 & 0.981 & 0.032

\tabularnewline
\hline
+ LUT \\ (1 iter)  & \textbf{34.692} & \textbf{0.981} & \textbf{0.024} & \underline{25.545}  & \underline{0.933} & \underline{0.075} & 33.977 & 0.982 &  0.022 & 37.237 & 0.982 & 0.031

\tabularnewline

+ LUT \\ (3 iters) &  \underline{34.568} & \underline{0.981} & \underline{0.022} &  25.522 & 0.932 & 0.076 & \underline{34.050} & \underline{0.983} &  \underline{0.021} & \textbf{37.263} & \textbf{0.981} & \textbf{0.031}
\tabularnewline

+ LUT \\ (5 iters)  & 34.544 & 0.981 & 0.024 & \textbf{25.574}  & \textbf{0.933} & \textbf{0.077} & \textbf{34.051}  & \textbf{0.983} &  \textbf{0.021} & \underline{37.241} & \underline{0.982} & \underline{0.032}

\tabularnewline
\hline

+ LUT \\ (1 basic)  & 34.587 & 0.980 & 0.026 &  \textbf{25.623} & \textbf{0.933}  & \textbf{0.075} & 33.794 & 0.982 & 0.023  & 37.006  & 0.981 & 0.033

\tabularnewline

+ LUT \\ (5 basics) &  \textbf{34.568} & \textbf{0.981} & \textbf{0.022} &  25.522 & 0.932 & 0.076 & \textbf{34.050} & \textbf{0.983} &  \textbf{0.021} & \underline{37.263} & \underline{0.981} & \underline{0.031}

\tabularnewline
+ LUT \\ (10 basics)  &  \underline{34.487} & \underline{0.981} & \underline{0.023} &  \underline{25.610} & \underline{0.933} & \underline{0.076} &  \underline{33.853} & \underline{0.983} & \underline{0.022}  & \textbf{37.350} & \textbf{0.982}  & \textbf{0.030}

\tabularnewline
\hline
\multirow{2}{*}{Method} & \multicolumn{3}{c}{{Lego}}  &  \multicolumn{3}{c}{{Materials}}  & \multicolumn{3}{c}{{Mics}} & \multicolumn{3}{c}{{Ship}} \\ 

 & PSNR $\uparrow$ & SSIM $\uparrow$ & LPIPS $\downarrow$ & PSNR $\uparrow$ & SSIM $\uparrow$ & LPIPS $\downarrow$  & PSNR $\uparrow$ & SSIM $\uparrow$ & LPIPS $\downarrow$  & PSNR $\uparrow$ & SSIM $\uparrow$ & LPIPS $\downarrow$

\tabularnewline
\hline

DVGO \cite{sun2021direct}  & 34.797  & 0.977 & 0.024 & 29.552  & 0.949 & 0.058 & 33.497 & 0.984 & 0.015  & 29.350 & 0.881 & 0.152

\tabularnewline
\hline
+ LUT \\ (w/o weight)  & 35.093 & 0.978 & 0.024 &  29.736 & 0.952 & 0.060 & 33.612  & 0.984 &  0.017 & 29.512 & 0.881 & 0.152

\tabularnewline
\hline
+ LUT \\ (1 iter)  & \textbf{35.435} & \textbf{0.979} & \textbf{0.023} &  29.952 & 0.954 & 0.056 & \underline{33.918} & \underline{0.985} & \underline{0.015}  & \textbf{29.726} & \textbf{0.883} & \textbf{0.151}

\tabularnewline
+ LUT \\ (3 iters) &  \underline{35.383}  & \underline{0.979} & \underline{0.023}  & \textbf{30.025}  & \textbf{0.955}  &  \textbf{0.055} & \textbf{33.931} & \textbf{0.985} & \textbf{0.015}  & \underline{29.545} & \underline{0.880} & \underline{0.152}

\tabularnewline

+ LUT \\ (5 iters)  & 35.257 & 0.978 & 0.024 & \underline{29.984}  & \underline{0.954} & \underline{0.055} & 33.861 & 0.986 & 0.0154  & 29.534 & 0.881 & 0.152

\tabularnewline
\hline

+ LUT \\ (1 basic)  & 35.131 & 0.978 & 0.024 & 29.898  & 0.953 & 0.056 & 33.733 & 0.985 & 0.016 & \textbf{29.465} & \textbf{0.882} & \textbf{0.150}

\tabularnewline
+ LUT \\ (5 basics) &  \textbf{35.383}  & \textbf{0.979} & \textbf{0.023}  & \textbf{30.025}  & \textbf{0.955}  &  \textbf{0.055} & \textbf{33.931} & \textbf{0.985} & \textbf{0.015}  & \underline{29.545} & \underline{0.880} & \underline{0.152}

\tabularnewline
+ LUT \\ (10 basics)  & \underline{35.370} & \underline{0.979} & \underline{0.023} &  \underline{29.943} & \underline{0.954} & \underline{0.056} & \underline{33.846} & \underline{0.985}  & \underline{0.016}  & 29.571 & 0.881 & 0.153

\tabularnewline

\hline

\end{tabular}
}

\label{t:lut ablation2}
\end{table*}

\begin{table*}[t]
\renewcommand\arraystretch{1.25}
\setlength{\belowcaptionskip}{0.1cm}
\centering
\caption{Ablation study of the LUTs refinement module in static scenes. We select the DVGO \cite{sun2021direct} and NeRF \cite{nerf} as the baseline method and evaluate 8 models in Synthetic-NeRF dataset \cite{nerf} and 4 models in TanksTemples dataset \cite{knapitsch2017tanks}. The number with bold typeface means the best.}
\setlength{\tabcolsep}{1.0mm}{}
\scalebox{0.75}{
\begin{tabular}{lccccccccc}

\hline

\multirow{2}{*}{Method} & \multicolumn{3}{c}{{Synthetic-NeRF}}  &  \multicolumn{3}{c}{{TanksTemples}} & \multicolumn{2}{c}{{Time}}  \\

 & PSNR $\uparrow$ & SSIM $\uparrow$ & LPIPS $\downarrow$ & PSNR $\uparrow$ & SSIM $\uparrow$ & LPIPS $\downarrow$ & train (h) $\downarrow$ & test (s) $\downarrow$

\tabularnewline
\hline

DVGO \cite{sun2021direct} & 32.135 & 0.957 & 0.050  & 28.739 & 0.908 & 0.162 & \textbf{0.31} & \textbf{0.366}

\tabularnewline
+ LUT  & \textbf{32.536} & \textbf{0.959} & \textbf{0.049} & \textbf{29.347} & \textbf{0.912} & \textbf{0.156} & 0.45 & 0.438

\tabularnewline
\hline

NeRF \cite{nerf} & 31.185 & 0.951 & 0.058 & 26.224 & 0.871 &  0.214  & \textbf{4.53} & \textbf{1.777}

\tabularnewline
+ LUT & \textbf{31.464} & \textbf{0.953}  & \textbf{0.056} & \textbf{26.512} & \textbf{0.872}  & \textbf{0.210} &  4.54 & 1.811

\tabularnewline
\hline

\end{tabular}
}

\label{t:lut ablation on 3d summary}
\end{table*}

\textbf{Ablation study on neural network architecture} Table \ref{t:v4d ablation} shows the ablation study on the design of V4D. First, we evaluate the design of the voxel and MLPs arrangement before the volume rendering. At first, we do the study on the two network variants which include the single voxel (SV) and single voxel-single radiance field (SF) formats as illustrated in Figure \ref{fig:svsf}. Comparing our full model (BL + (D, C, L)) with SV and SF, we can see that the proposed V4D in dual voxel setting and modeling the density and texture fields separately could achieve a better result. We can learn that the performance in the SF setting is worse in some subsets such as in \emph{Bouncing Balls} and \emph{T-Rex}. The potential explanation behind this phenomenon is that sharing the same MLPs network for the density and RGB value would cause some contradiction when optimizing the voxel grid. As stated before, at the beginning of the training, the density should be initialized into zero for the correct volume rendering, and sharing too many features with RGB values may make the training unstable. Second, we evaluate the contributions of different components in our network. We can see that the decay total variation loss working on the voxel grid is necessary for the training, which could prevent the model from overfitting to the training set. However, the total variation loss tends to cause over smooth visual performance. From Figure \ref{flip on 4d}, we can see that the appearance is a bit blurred even though we decrease the weight of the total variation loss with the exponential decay strategy during the training phase. To handle this problem, the proposed conditional positional encoding (CPE) and LUTs refinement module could alleviate the problem of over-smoothness and the full setting of our method achieves generally better results, which verified the effectiveness. For a better visual analysis, we select several results in the format of the FLIP error map \cite{flip} in Figure \ref{flip on 4d}.

\textbf{Hyperparameters on the LUTs refinement module}
The hyperparameters ablation study of the LUTs refinement module is summarized in Table \ref{t:lut ablation2}. For the experiment, we train 100k iterations for all the settings. The LUT refinement module works at the beginning of the training. The iter/iters setting means the times of the recurrent iteration and the basic/basics setting defines the number of the basic LUT in the refinement module. We can learn that the experiment LUT refinement module setting with 3 iterations and 5 basic LUTs achieves the best result. The 3 iterations setting is slightly better than the 1 iteration, but with more iterations in 5, it could degrade the performance. About the number of basic LUT, 1 basic LUT can achieve performance improvement compared with the baseline, and using 5 basic LUTs could achieve a better result with more expressive color representation ability composed of different basic LUTs. We did not observe further improvement by increasing the basic LUT to 10. Therefore, we choose the LUTs refinement module with 3 recurrent iterations and 5 basic LUTs units. Note that the experiment without the spatial awareness weight from the pseudo-surface is generally worse than the proposed setting, which shows that the pseudo-surface could guide the refinement of learning with the local 3D information.

\begin{figure*}[t]
	\setlength{\belowcaptionskip}{-0.3cm}
	\centering
	\begin{minipage}[t]{0.03\linewidth}
	    \rotatebox{90}{\quad DVGO}
	\end{minipage}
	\begin{minipage}[t]{0.102\linewidth}
		\centerline{\includegraphics[height=1.452cm]{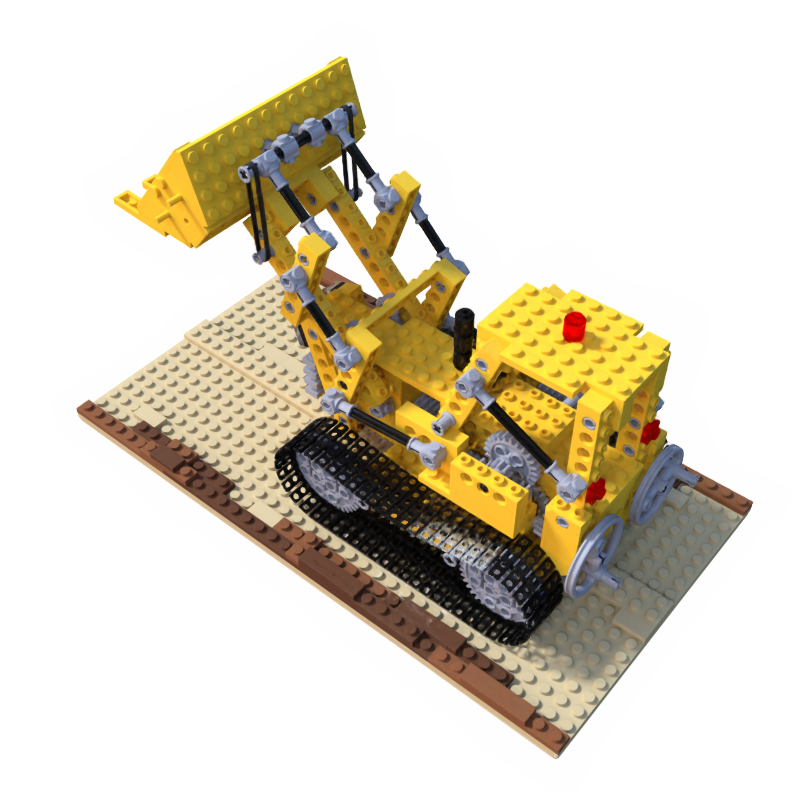}}
	\end{minipage}
	\begin{minipage}[t]{0.102\linewidth}
		\centerline{\includegraphics[height=1.452cm]{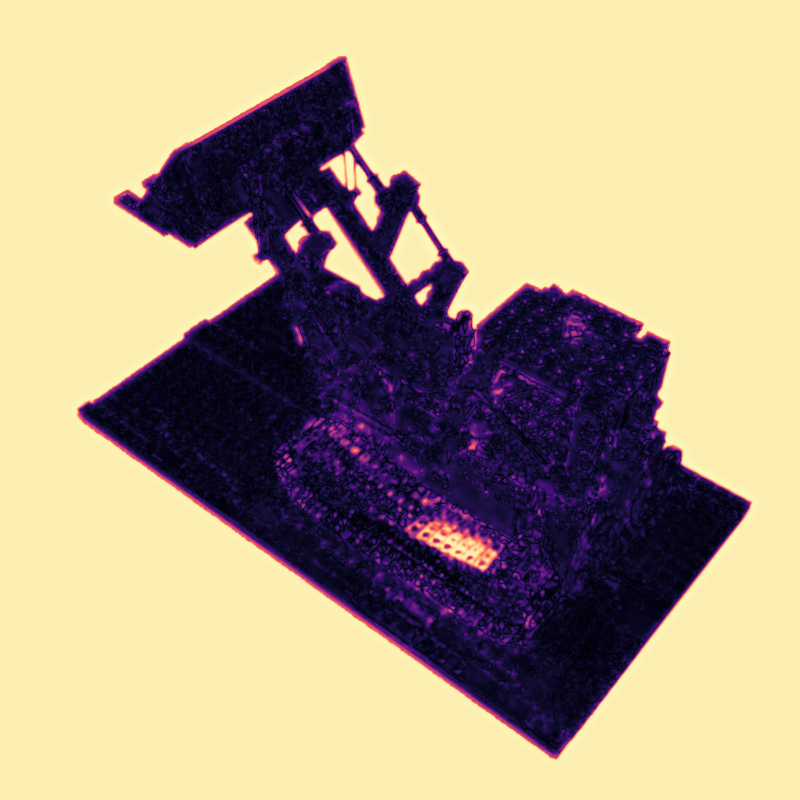}}
	\end{minipage}
	\begin{minipage}[t]{0.182\linewidth}
		\centerline{\includegraphics[height=1.452cm]{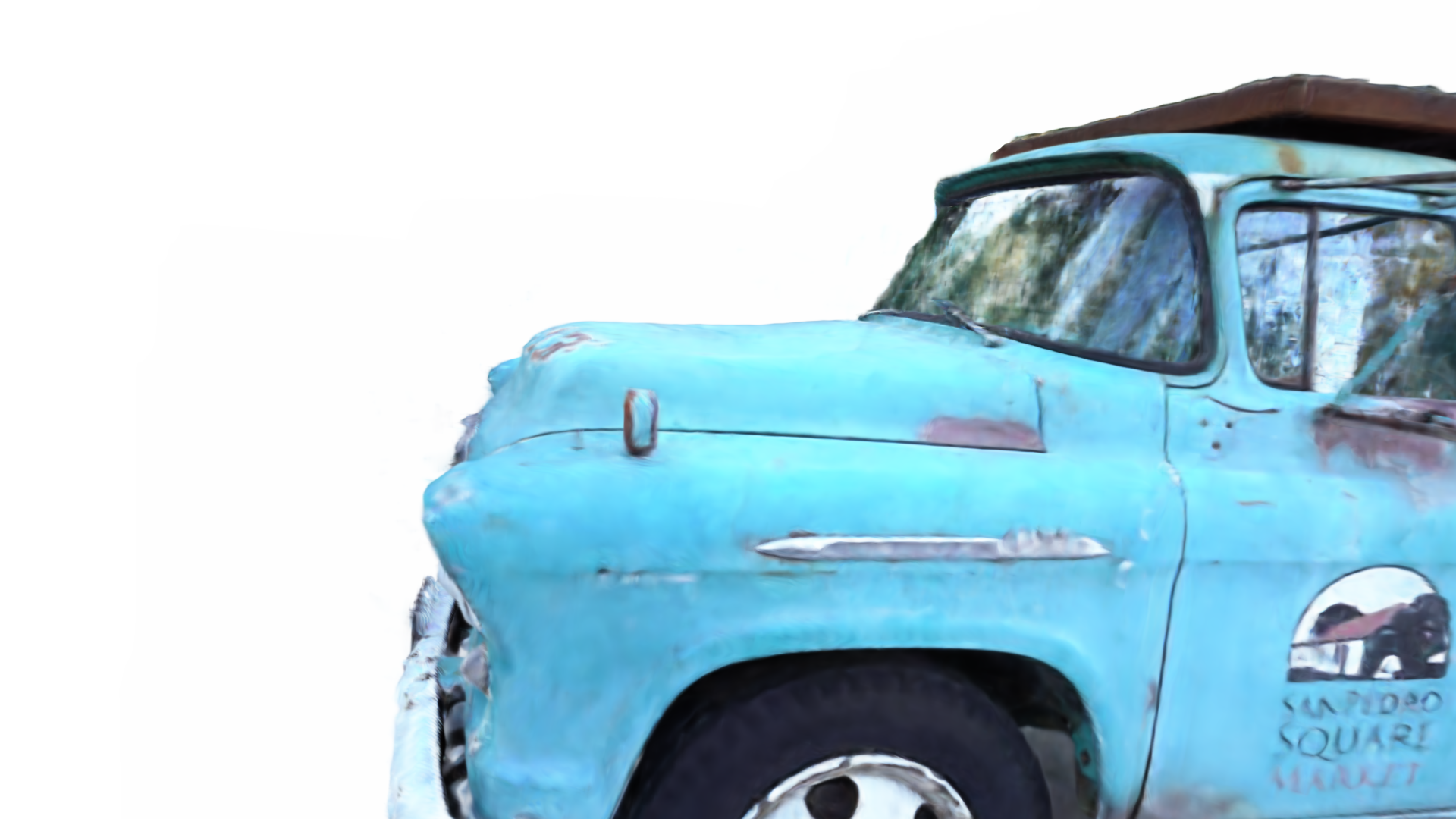}}
	\end{minipage}
	\begin{minipage}[t]{0.182\linewidth}
		\centerline{\includegraphics[height=1.452cm]{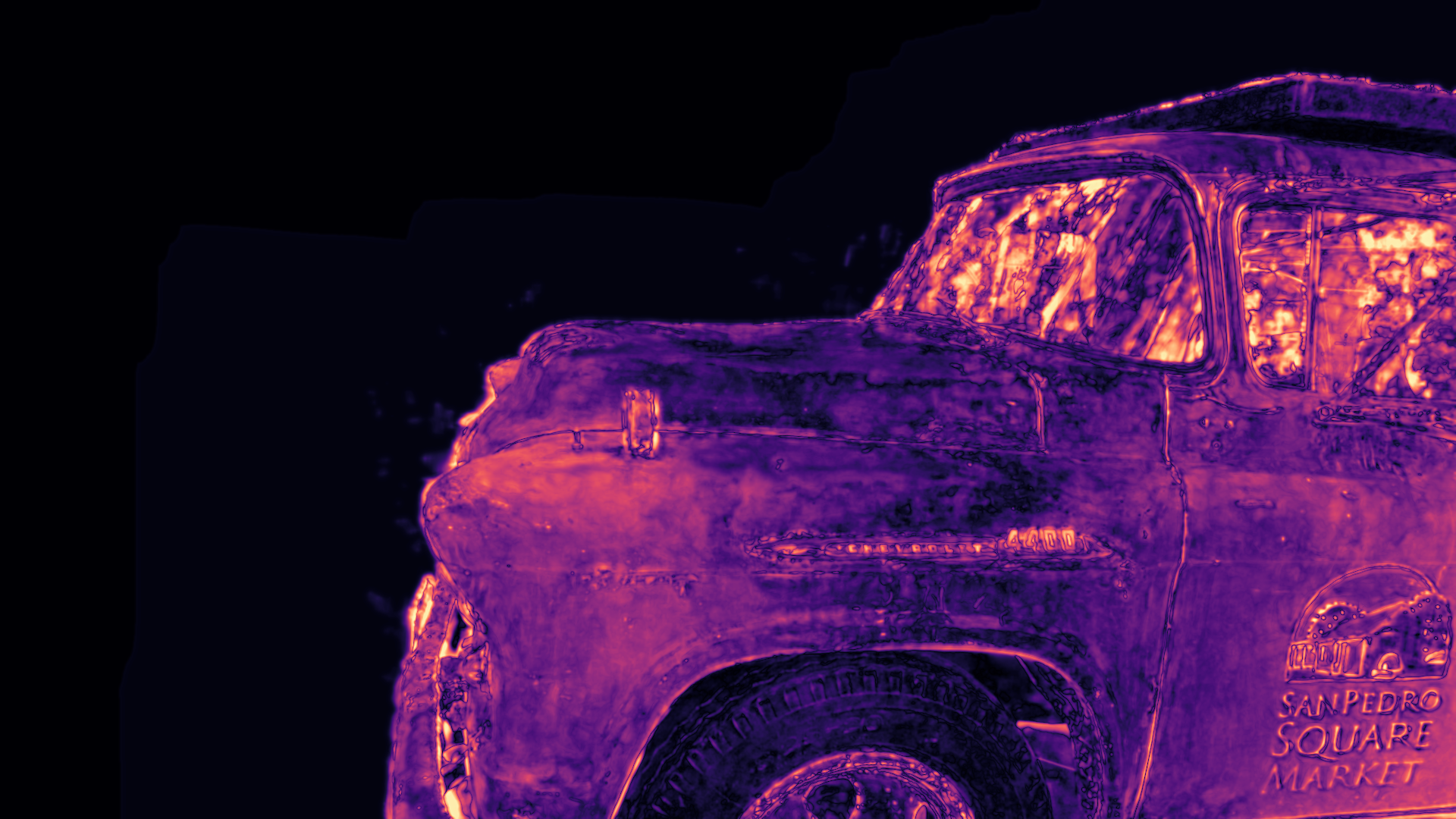}}
	\end{minipage}
	\begin{minipage}[t]{0.182\linewidth}
		\centerline{\includegraphics[height=1.452cm]{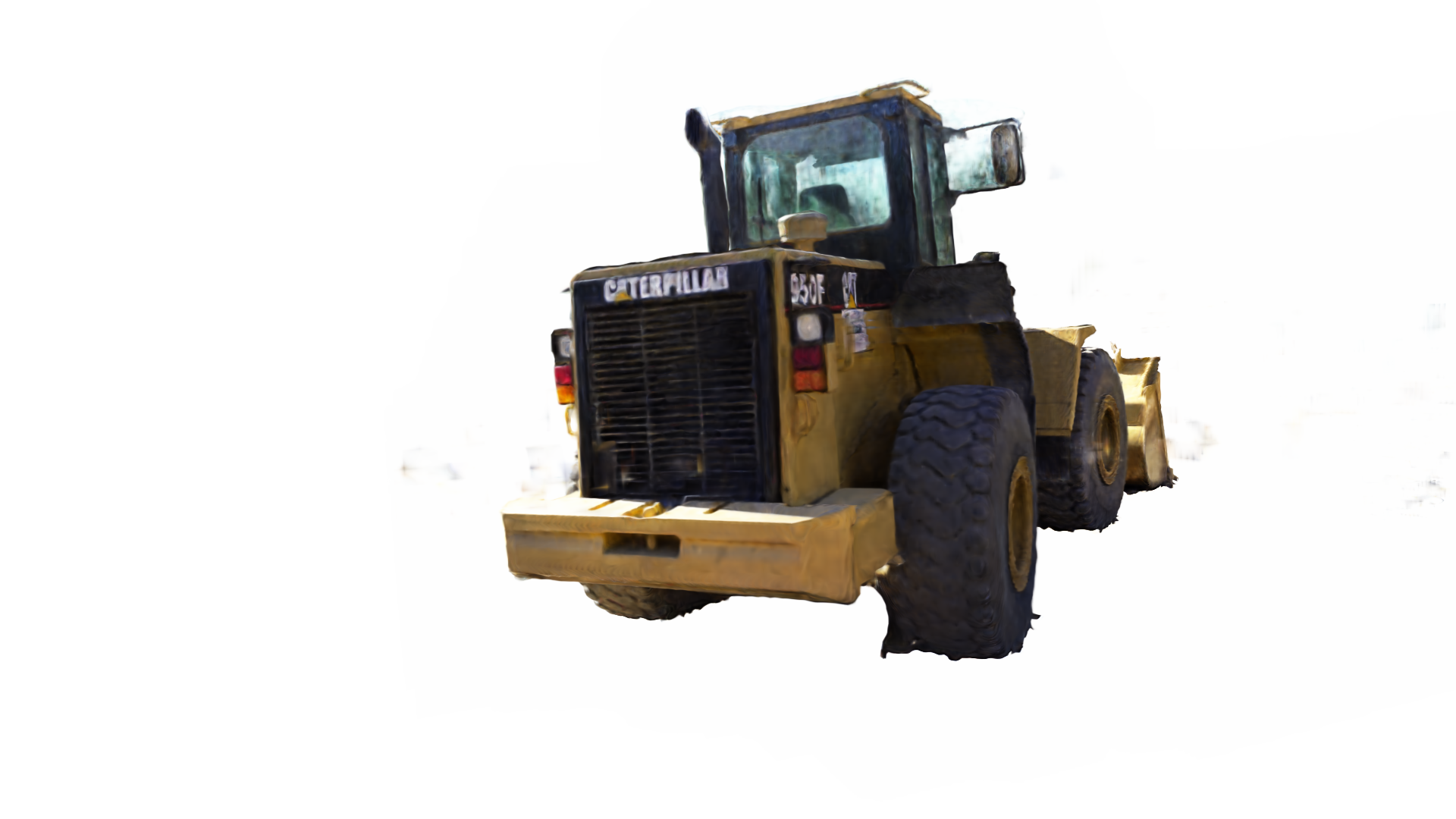}}
	\end{minipage}
	\begin{minipage}[t]{0.182\linewidth}
		\centerline{\includegraphics[height=1.452cm]{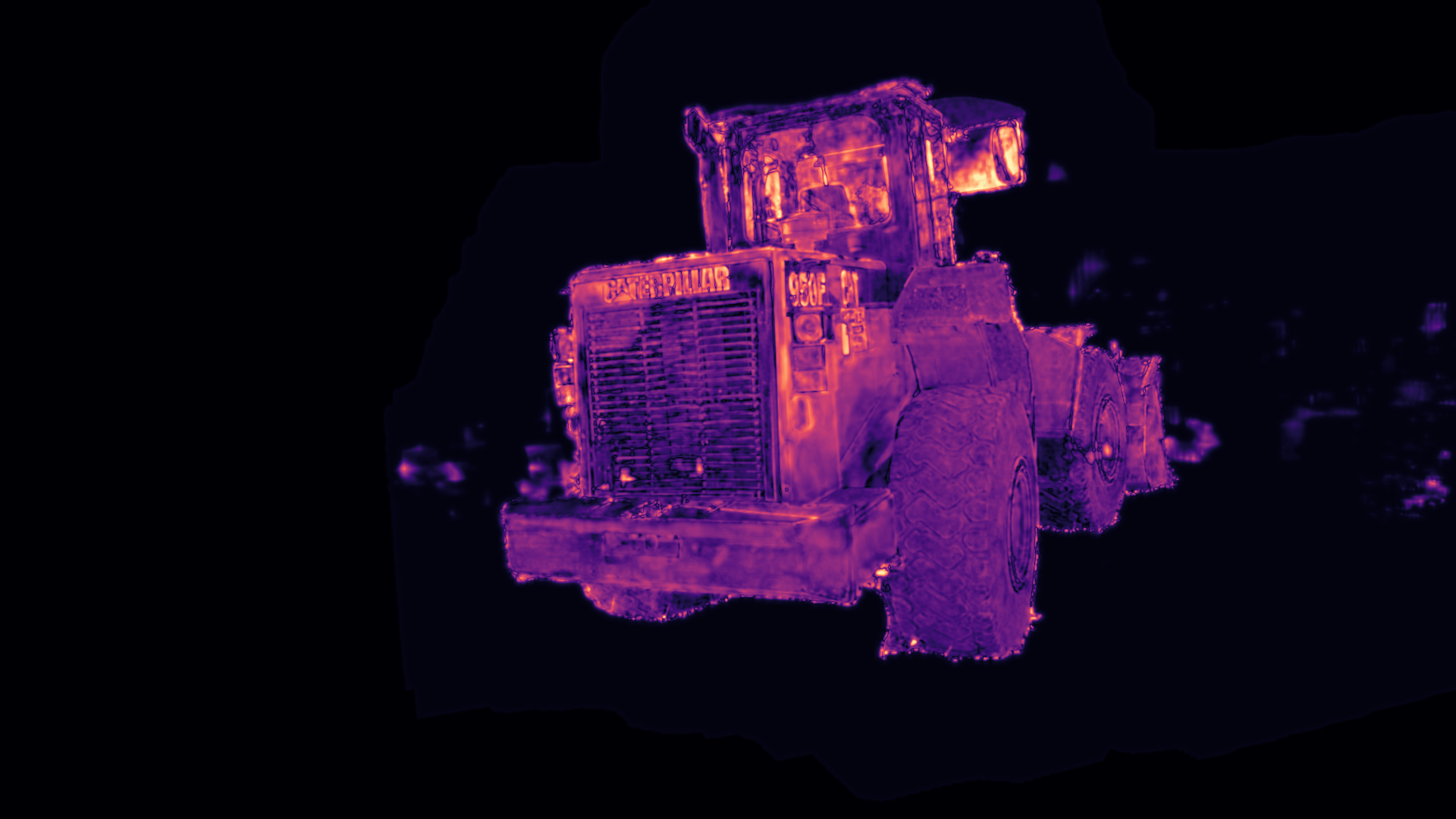}}
	\end{minipage}
	\hfill
	\begin{minipage}[t]{0.03\linewidth}
	    \rotatebox{90}{DVGO+L}
	\end{minipage}
	\begin{minipage}[t]{0.102\linewidth}
		\centerline{\includegraphics[height=1.452cm]{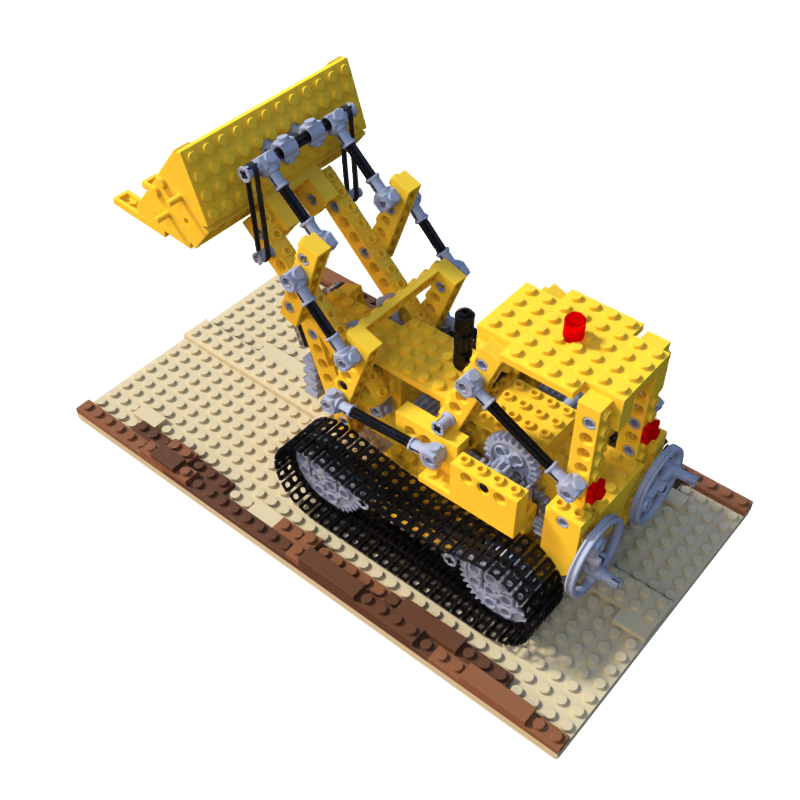}}
		\centerline{Lego}
	\end{minipage}
	\begin{minipage}[t]{0.102\linewidth}
		\centerline{\includegraphics[height=1.452cm]{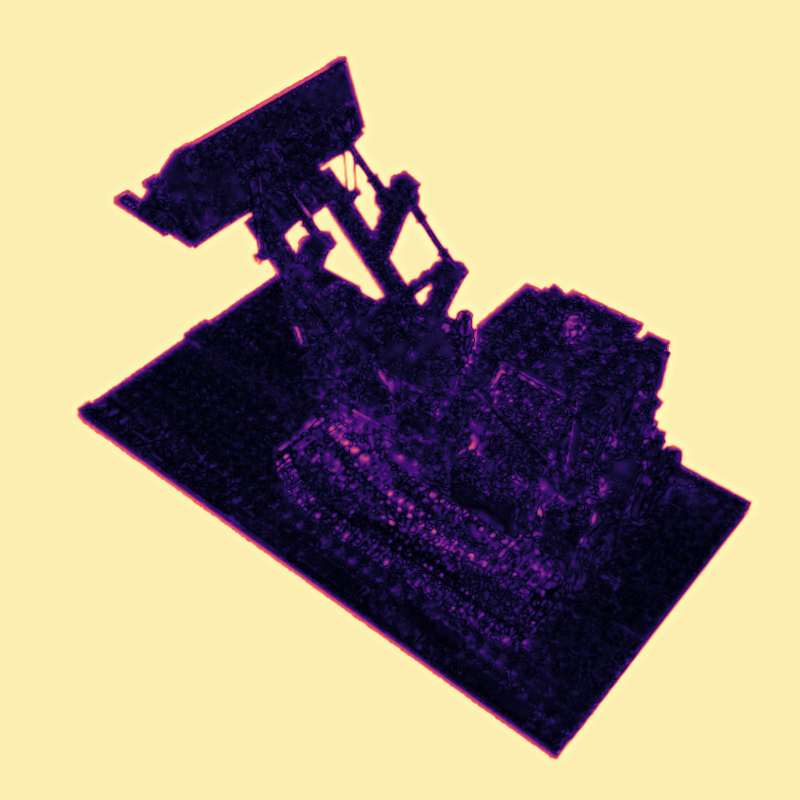}}
	\end{minipage}
	\begin{minipage}[t]{0.182\linewidth}
		\centerline{\includegraphics[height=1.452cm]{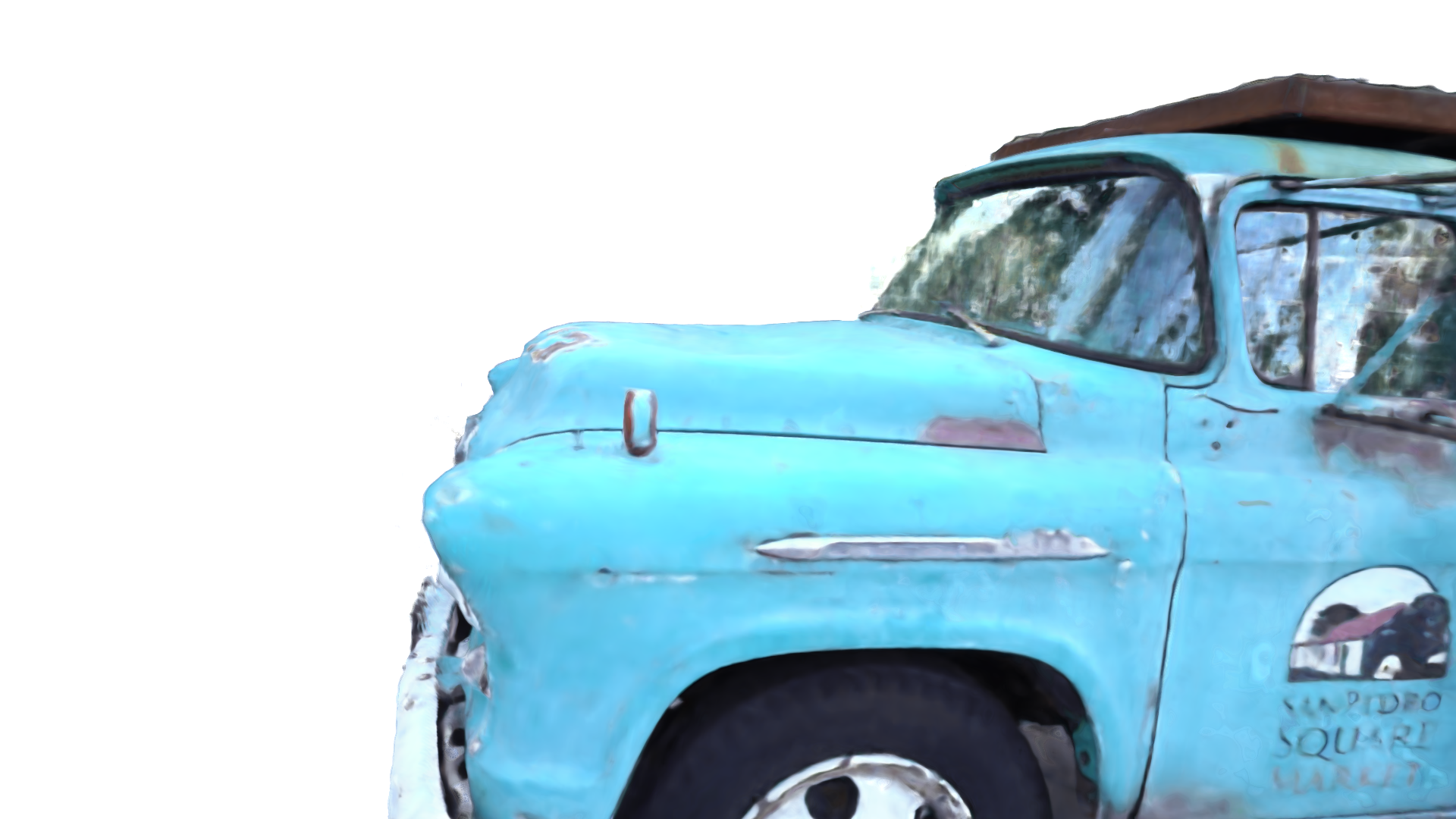}}
		\centerline{Truck}
	\end{minipage}
	\begin{minipage}[t]{0.182\linewidth}
		\centerline{\includegraphics[height=1.452cm]{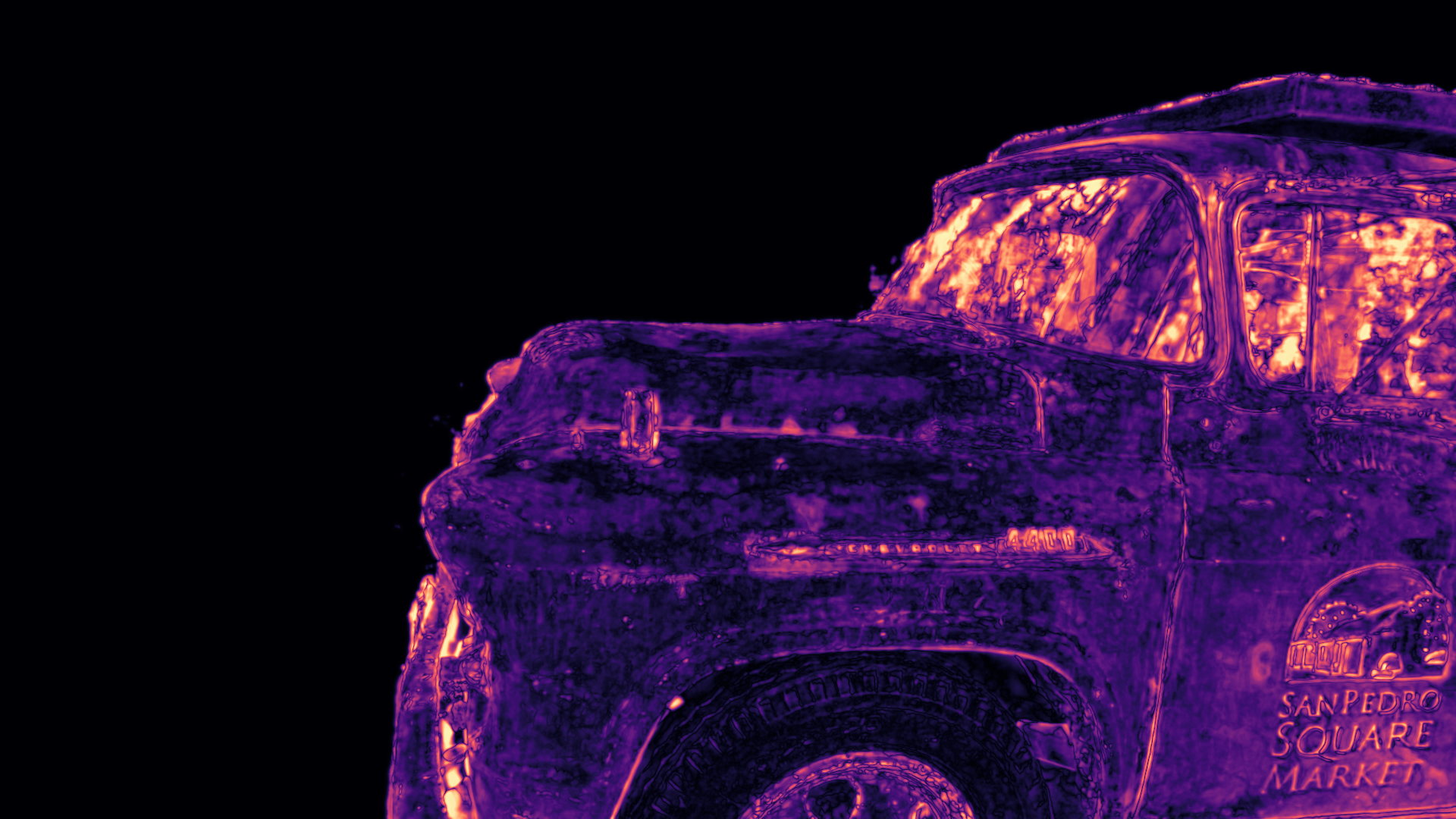}}
	\end{minipage}
	\begin{minipage}[t]{0.182\linewidth}
		\centerline{\includegraphics[height=1.452cm]{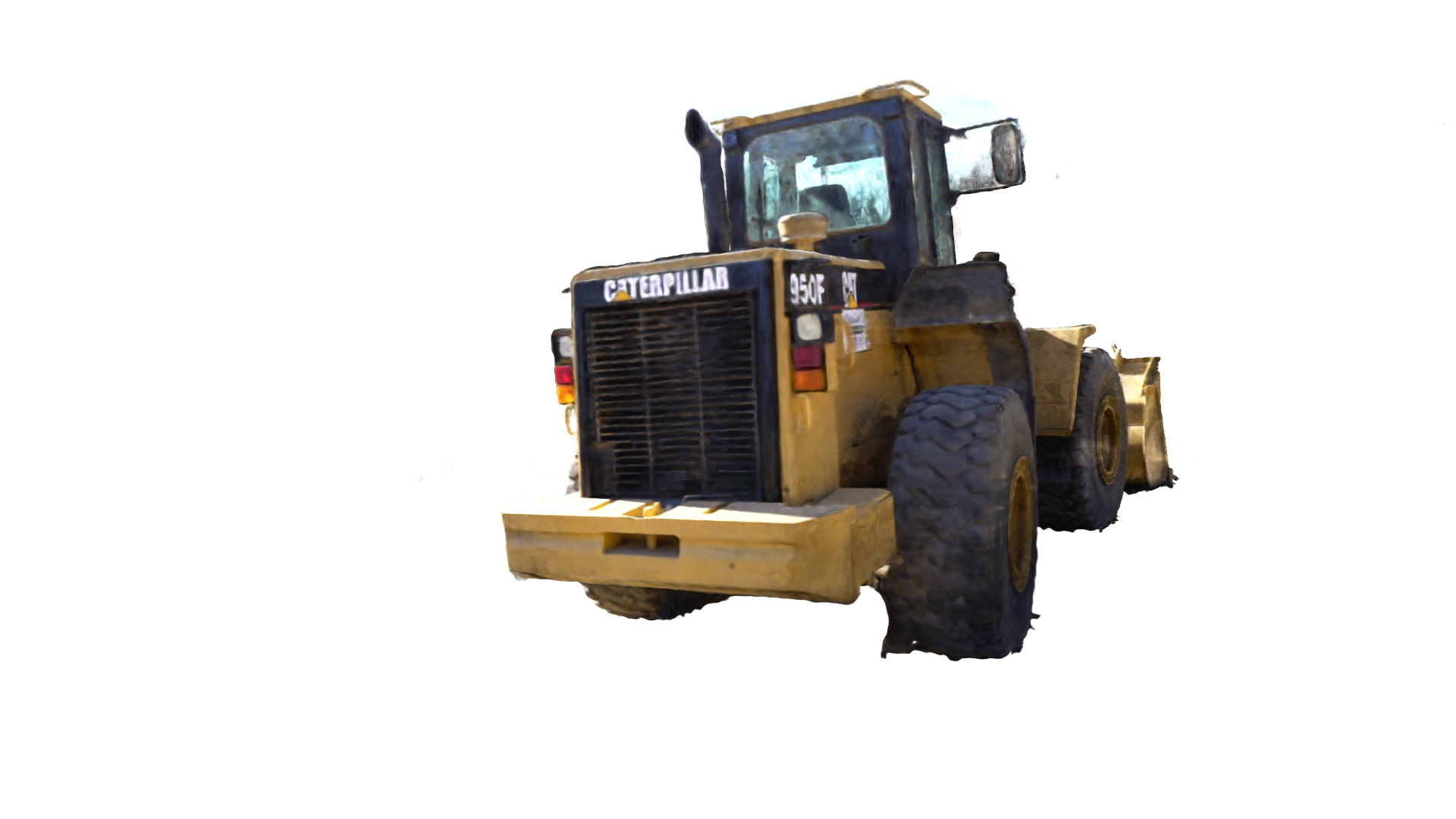}}
		\centerline{Caterpillar}
	\end{minipage}
	\begin{minipage}[t]{0.182\linewidth}
		\centerline{\includegraphics[height=1.452cm]{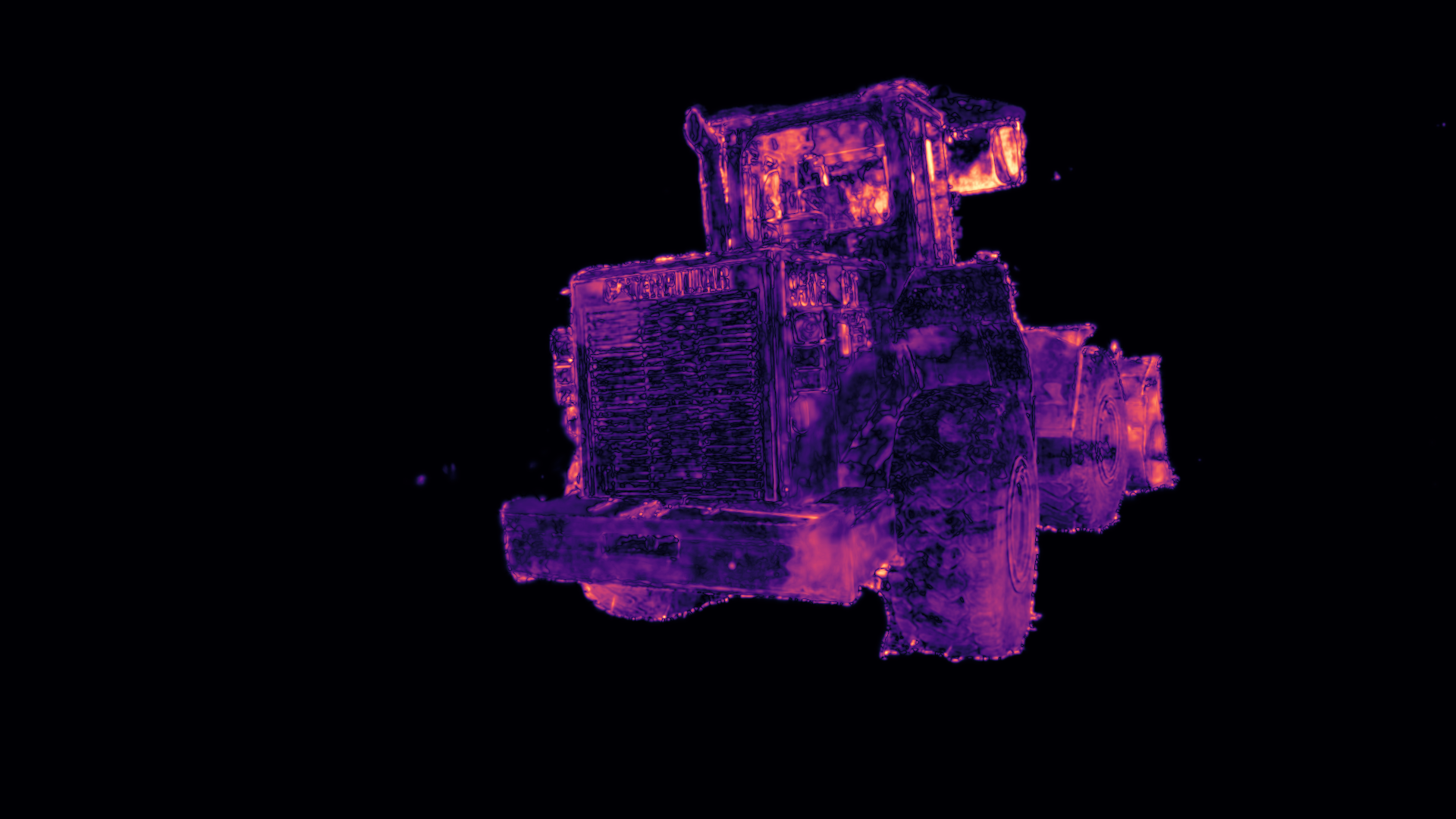}}
	\end{minipage}
	\centering
	\caption{FLIP error map \cite{flip} visualization on the ablation studies about the LUTs refinement module in 3D novel view synthesis. Brighter means a larger error. Please zoom in for better observation.}\label{FLIP in 3d}
\end{figure*}

\textbf{LUTs refinement module in 3D novel view synthesis}
The proposed LUTs refinement module is also suitable for the 3D novel view synthesis, which is a plug-and-play module in the novel view synthesis task. We choose the DVGO \cite{sun2021direct} and NeRF \cite{nerf} as the baseline. For DVGO \cite{sun2021direct}, we train 100k iterations for all the 3D datasets for a fair comparison. For NeRF \cite{nerf}, we only train 50k iterations and discard the coarse sampling deal to the limitation of the computational resource. From Table \ref{t:lut ablation on 3d summary}, we can see that the improvement is obvious compared with the baseline. Besides, the computational cost with LUTs refinement is acceptable thanks to the refinement in the pixel level. Note that the overall performance improvement is better than the result in Table \ref{t:v4d ablation} and the reason is that the geometry of the pseudo-surface in the static dataset is better than in the dynamic dataset, which could offer more precise guidance in the LUTs' weight prediction. For a better visual analysis, we select several results of this ablation study in the format of the FLIP error map \cite{flip} in Figure \ref{FLIP in 3d}. A more detailed study of the LUTs refinement module and the full numerical result are presented in the supplemental material.

\subsection{Limitation and future work} \label{limitation}
We achieve excellent performance with the voxel-based architecture for the 4D novel view synthesis task, but one obvious limitation is that the voxel format requires larger memory storage compared with the pure MLPs-based method (e.g., V4D (377 MB) versus D-NeRF (13 MB)), which means that it is limited to the resolution and not very suitable for the large scale 4D scene representation. However, the recent work in \cite{chen2022tensorf} proposes to factorize the 4D voxel into multiple compact low-rank tensor components that could close the gap between the voxel-based method and pure MLPs-based method in the memory storage aspect. At last, we believe the proposed framework should be also suitable for the dynamic reconstruction task \cite{cai2022neural} with the feature without the need for the canonical space, which would be considered as future work.

\section{Conclusion}
In this paper, we present a new framework, V4D, for 4D novel view synthesis. The voxel-based framework could effectively overcome the limited capacity and the high computational cost problems in pure MLPs-based methods, which shows significant improvement and achieves state-of-the-art performance. The proposed conditional positional encoding and the LUTs refinement module also benefit further improvement, especially for alleviating the problem of over-smoothness caused by the total variation loss. The voxel-based LUTs refinement module could be regarded as the plug-and-play module in the novel view synthesis task, which could achieve the performance gain at little computational cost. We hope the LUTs refinement module gives some inspiration for the follower on the refinement angle in the novel view synthesis task.

\clearpage

\bibliographystyle{plain}

\bibliography{ref}

\end{document}